\documentclass[10pt,letterpaper,twocolumn]{article}
\usepackage[latin9]{inputenc}
\usepackage{array}
\usepackage{multirow}
\usepackage{amsmath}
\usepackage{amssymb}
\usepackage{graphicx}
\usepackage{wasysym}
\usepackage[unicode=true,
 bookmarks=false,
 breaklinks=true,pdfborder={0 0 1},backref=section,colorlinks=false]
 {hyperref}
\hypersetup{
 pagebackref=true,letterpaper=true,colorlinks}

\makeatletter

\pdfpageheight\paperheight
\pdfpagewidth\paperwidth

\providecommand{\tabularnewline}{\\}

\usepackage{cvpr}
\usepackage{times}
\usepackage{epsfig}
\usepackage{graphicx}

\cvprfinalcopy 

\def\cvprPaperID{****} 

\ifcvprfinal\fi

\makeatother

\begin{document}
\title{SM-NAS: Structural-to-Modular Neural Architecture Search for Object
Detection}
\author{Lewei Yao$^{1*}$\hspace{4mm}Hang Xu$^{1}$\thanks{Equal contribution}
\hspace{4mm}Wei Zhang$^{1}$\hspace{4mm}Xiaodan Liang$^{2}$\thanks{Corresponding author:\textit{ }xdliang328@gmail.com}
\hspace{4mm}Zhenguo Li$^{1}$\\
$^{1}$Huawei Noah's Ark Lab \hspace{5mm}$^{2}$Sun Yat-sen University}
\maketitle
\begin{abstract}
The state-of-the-art object detection method is complicated with various
modules such as backbone, feature fusion neck, RPN, and RCNN head,
where each module may have different designs and structures. How to
leverage the computational cost and accuracy trade-off for the structural
combination as well as the modular selection of multiple modules?
Neural architecture search (NAS) has shown great potential in finding
an optimal solution. Existing NAS works for object detection only
focus on searching better design of a single module such as backbone
or feature fusion neck, while neglecting the balance of the whole
system. In this paper, we present a two-stage coarse-to-fine searching
strategy named Structural-to-Modular NAS (SM-NAS) for searching a
GPU-friendly design of both an efficient combination of modules and
better modular-level architecture for object detection. Specifically,
Structural-level searching stage first aims to find an efficient combination
of different modules; Modular-level searching stage then evolves each
specific module and pushes the Pareto front forward to a faster task-specific
network. We consider a multi-objective search where the search space
covers many popular designs of detection methods. We directly search
a detection backbone without pre-trained models or any proxy task
by exploring a fast training from scratch strategy. The resulting
architectures dominate state-of-the-art object detection systems in
both inference time and accuracy and demonstrate the effectiveness
on multiple detection datasets, e.g. halving the inference time with
additional 1\% mAP improvement compared to FPN and reaching 46\% mAP
with the similar inference time of MaskRCNN.
\end{abstract}

\section{Introduction}

\begin{figure}[th]
\begin{centering}
\includegraphics[scale=0.5]{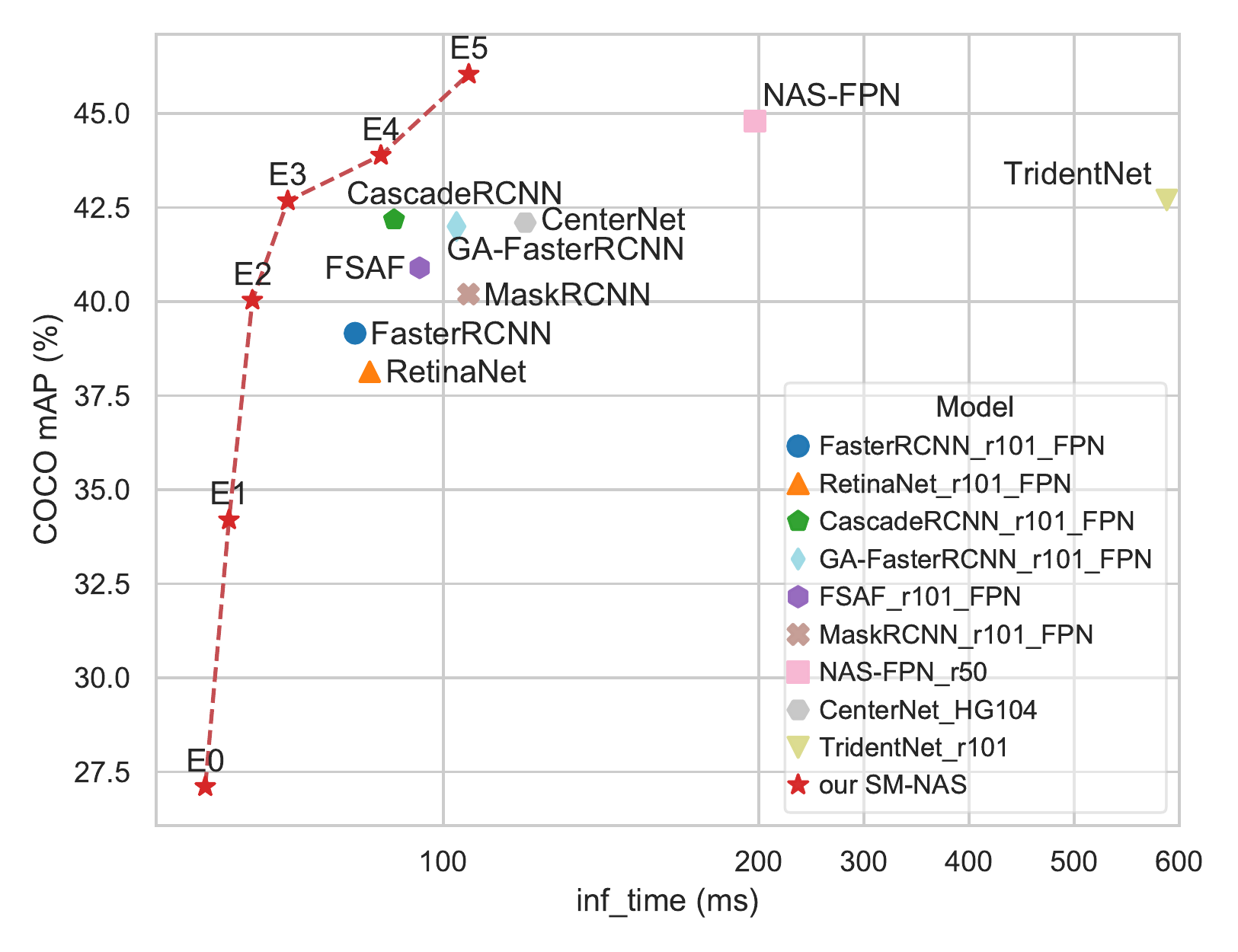}
\par\end{centering}
\vspace{-2mm}

\caption{\label{fig:benchmark-graph}Inference time (ms) and detection accuracy
(mAP) comparison on COCO dataset. SM-NAS yields state-of-the-art speed/accuracy
trade-off. The compared models include our networks (E0 to E5) and
classical detectors (Faster-RCNN w FPN \cite{lin2017feature}, RetinaNet
\cite{he2016deep}) and most recent works (TridentNet\cite{Li2019}
, NAS-FPN\cite{Ghiasi_2019_CVPR}, etc.).}

\vspace{-2mm}
\end{figure}
Real-time object detection is a core and challenging task to localize
and recognize objects in an image on a certain device. This task widely
benefits autonomous driving \cite{chabot2017deep}, surveillance video
\cite{luo2014switchable}, facial recognition in mobile phone \cite{bhagavatula2017faster},
to name a few. A state-of-the-art detection system \cite{liu2016ssd,redmon2016you,ren2015faster,lin2017feature,he2018rethinking,lin2017focal}
usually consists of four modules: backbone, feature fusion neck, region
proposal network (in two-stage detection), and RCNN head. Recent progress
in this area shows various designs of each modules: backbone \cite{he2016deep,redmon2017yolo9000,li2018detnet},
region proposal network\cite{wang2019region}, feature fusion neck
\cite{lin2017feature,liu2018path,li2019scale} and RCNN head \cite{dai2016r,li2017light,cai2017cascade}.

However, how to select the best combination of modules under hardware
resource constrains remains unknown. This problem draws much attention
from the industry because in practice adjusting each module manually
based on a standard detection model is inefficient and sub-optimal.
It is hard to leverage and evaluate the inference time and accuracy
trade-off as well as the representation capacity of each module in
different datasets. For instance, empirically we found that combination
of Cascade-RCNN with ResNet18 (not a standard detection model) is
even faster and more accurate than FPN with ResNet50 in COCO \cite{lin2014microsoft}
and BDD \cite{yu2018bdd100k} (autonomous driving dataset). However,
this is not true in the case of VOC\cite{Everingham10}.

There has been a growing trend in automatically designing a neural
network architecture instead of relying heavily on human efforts and
experience. For the image classification, \cite{zoph2018learning,liu2018progressive,liu2018darts,real2018regularized,tan2019efficientnet}
searched networks surpass the performance of hand-crafted networks.
For the detection task, existing NAS works focus on optimizing a single
component of the detection system instead of considering the whole
system. For example, \cite{zoph2018learning} only transfers the searched
architecture from the classification task (ImageNet) to the detector
backbone. DetNAS\cite{chen2019detnas} searches for better backbones
on a pre-trained super-net for object detection. NAS-FPN\cite{Ghiasi_2019_CVPR},
Auto-FPN\cite{Xu_2019_ICCV}, NAS-FCOS\cite{wang2019fcos} use NAS
to find a better feature fusion neck and a more powerful RCNN head.
However, those pipelines only partially solve the problem by changing
one component while neglecting the balance and efficiency of the whole
system. On the contrary, our work aims to develop a multi-objective
NAS scheme specifically designed to find an optimal and efficient
whole architecture.In this work, we make the first effort on searching
the whole structure for object detectors. By investigating the state-of-the-art
design, we found three factors are crucial for the performance of
a detection system: 1) size of the input images; 2) combination of
modules of the detector; 3) architecture within each module. To find
an optimal tradeoff between inference time and accuracy with these
three factors, we propose a coarse-to-fine searching strategy: 1)
Structural-level searching stage (Stage-one) first aims to find an
efficient combination of different modules as well as the model-matching
input sizes; 2) Modular-level search stage (Stage-two) then evolves
each specific module and push forward to an efficient task-specific
network.

We consider a multi-objective search targeting directly on GPU devices,
which outputs a Pareto front showing the optimal designs of the detector
under different resource constraints. During Stage-one, the search
space includes different choices of modules to cover many popular
one-stage/two-stage designs of detectors. We also consider putting
the input image size into the search space since it greatly impacts
the latency and accuracy \cite{tan2019efficientnet}. During Stage-two,
we further consider to optimize and evolve the modules (e.g. backbone)
following the optimal combination found in the previous stage. The
previous works \cite{li2018detnet} find that backbones originally
designed for classification task might be sub-optimal for object detection.
The resulting modular-level search thus leans the width and depth
of the overall architecture towards detection task. With the improved
training strategy, our search can be conducted directly on the detection
datasets without ImageNet pre-training. For an efficient search, we
combine evolutionary algorithms \cite{real2017large,real2018regularized}
with Partial Order Pruning technique \cite{li2019partial} for a fast
searching and parallelize the whole searching algorithm in a distributed
training system to further speed up the whole process.

Extensive experiments are conducted on the widely used detection benchmarks,
including Pascal VOC \cite{Everingham10}, COCO \cite{lin2014microsoft},
BDD \cite{yu2018bdd100k}. As shown in Figure \ref{fig:benchmark-graph},
SM-NAS yields state-of-the-art speed/accuracy trade-off and outperforms
existing detection methods, including FPN \cite{lin2017feature},
Cascade-RCNN \cite{cai2017cascade} and the most recent work NAS-FPN
\cite{Ghiasi_2019_CVPR}. Our E2 reaches half of the inference time
with additional 1\% mAP improvement compared to FPN. E5 reaches 46\%
mAP with the similar inference time of MaskRCNN (mAP:39.4\%).

To sum up, we make the following contributions to NAS for detection:
\begin{itemize}
\item We are among the first to investigate the trade-off for speed and
accuracy of an object detection system with a different combination
of different modules.
\item We develop a coarse-to-fine searching strategy by decoupling the search
into structural-level and modular-level to efficiently lift the Pareto
front. The searched models reach the state-of-the-art speed/accuracy,
dominating existing methods with a large margin.
\item We make the first attempt to directly search a detection backbone
without pre-trained models or any proxy task by exploring fast training
from scratch strategy.
\end{itemize}

\section{Related Work}

\begin{figure*}
\begin{centering}
\vspace{-2mm}
\includegraphics[scale=0.55]{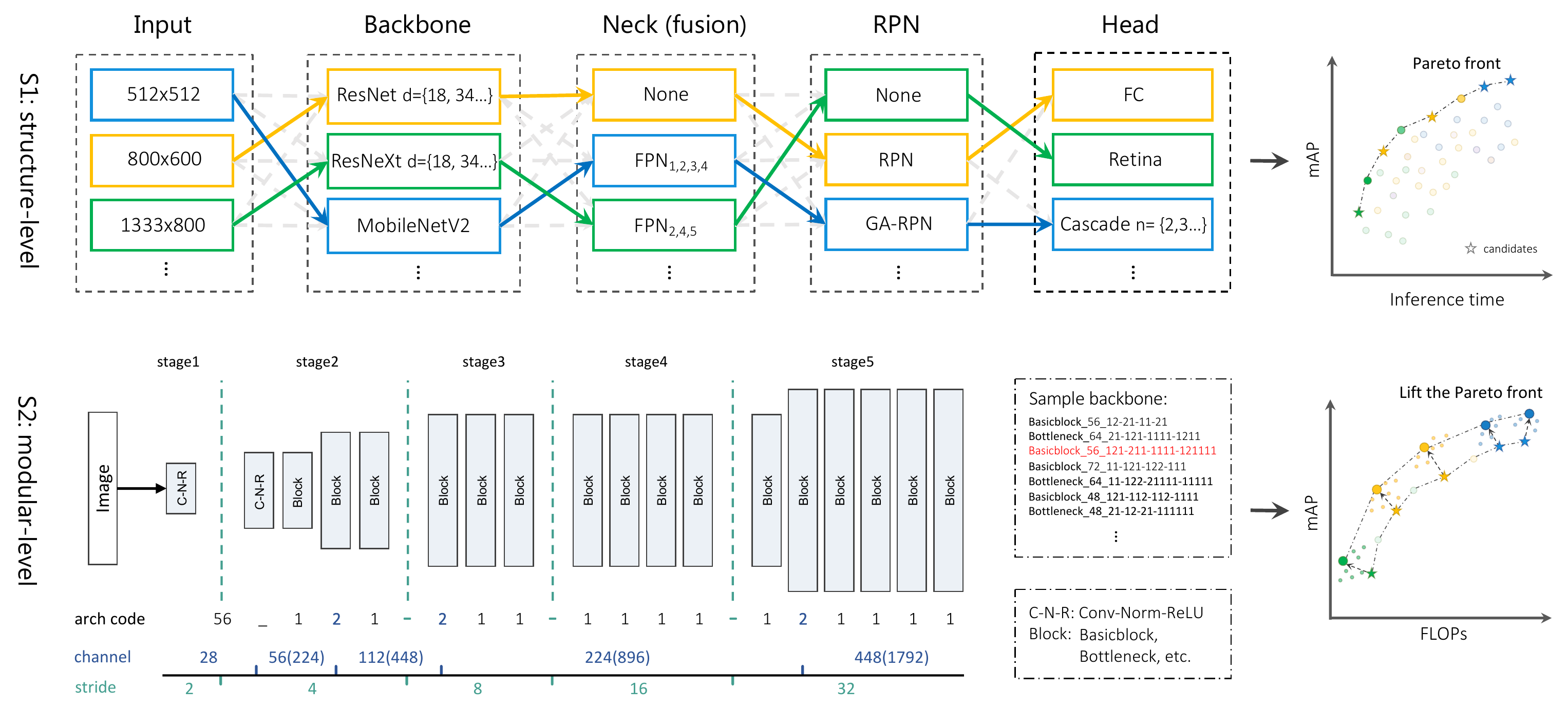}
\par\end{centering}
\vspace{-2mm}

\caption{\label{fig:framework-graph} An overview of our SM-NAS for detection
pipeline. We propose a two-stage coarse-to-fine searching strategy
directly on detection dataset: S1: Structural-level searching stage
first aims to finding an efficient combination of different modules;
S2: Modular-level search stage then evolves each specific module and
push forward to a faster task-specific network.}

\vspace{-2mm}
\end{figure*}

\textbf{Object Detection.} Object detection is a core problem in computer
vision. State-of-the-art anchor-based detection approaches usually
consists of four modules: backbone, feature fusion neck, region proposal
network (in two-stage detectors), and RCNN head. Most of the previous
progress focus on developing better architectures for each module.
For example, \cite{li2018detnet} tries to develop a backbone for
detection; FPN \cite{lin2017feature} and PANet \cite{liu2018path}
modified multi-level features fusion module; \cite{wang2019region}
try to make RPN more powerful. On the other hand, R-FCN \cite{dai2016r}
and Light-head RCNN \cite{li2017light} design different structures
of bbox head. However, community lacks of literatures comparing the
efficiency and performance of different combination of different modules.

\textbf{Neural Architecture Search. }NAS aims at automatically finding
an efficient neural network architecture for a certain task and dataset
without labor of designing network. Most works are based on searching
CNN architectures for image classification while only a few of them
\cite{chen2018searching,liu2019auto,chen2019detnas} focus on more
complicated vision tasks such as semantic segmentation and detection.
There are mainly three categories of searching strategies in NAS area:
1) Reinforcement learning based methods \cite{baker2016designing,zoph2018learning,cai2018efficient,zhong2018practical}
train a RNN policy controller to generate a sequence of actions to
specify CNN architecture; 2) Evolutionary Algorithms based methods
and Network Morphism \cite{real2017large,liu2017hierarchical,real2018regularized}
try to \textquotedblleft evolves\textquotedblright{} architectures
by mutating the current best architectures; 3) Gradient based methods
\cite{liu2018darts,xie2018snas,cai2018proxylessnas} define an architecture
parameter for continuous relaxation of the discrete search space,
thus allowing differentiable optimization of the architecture. Among
those approaches, gradient based methods is fast but not so reliable
since weight-sharing makes a big gap between the searching and final
training. RL methods usually require massive samples to converge which
is not practical for detection. Thus we use EA based method in this
paper.

\vspace{-2mm}

\section{The Proposed Approach}

\subsection{Motivation and preliminary experiments}

With preliminary empirical experiments, we have found some interesting
facts:

\begin{table}
\begin{centering}
\tabcolsep 0.02in{\footnotesize{}}%
\begin{tabular}{ccccccc}
\hline 
{\footnotesize{}id} & {\footnotesize{}Dataset} & {\footnotesize{}Model} & {\footnotesize{}Backbone} & {\footnotesize{}Input Img} & {\footnotesize{}Time(ms)} & {\footnotesize{}mAP}\tabularnewline
\hline 
{\footnotesize{}1} & {\footnotesize{}COCO} & {\footnotesize{}FPN} & {\footnotesize{}ResNet50} & {\footnotesize{}800x600} & {\footnotesize{}43.6} & {\footnotesize{}36.3}\tabularnewline
{\footnotesize{}2} & {\footnotesize{}COCO} & {\footnotesize{}RetinaNet} & {\footnotesize{}ResNet50} & {\footnotesize{}800x600} & {\footnotesize{}46.7} & {\footnotesize{}34.8}\tabularnewline
{\footnotesize{}3} & {\footnotesize{}VOC} & {\footnotesize{}FPN} & {\footnotesize{}ResNet50} & {\footnotesize{}800x600} & {\footnotesize{}38.4} & {\footnotesize{}80.4}\tabularnewline
{\footnotesize{}4} & {\footnotesize{}VOC} & {\footnotesize{}RetinaNet} & {\footnotesize{}ResNet50} & {\footnotesize{}800x600} & {\footnotesize{}34.8} & {\footnotesize{}79.7}\tabularnewline
{\footnotesize{}5} & {\footnotesize{}COCO} & {\footnotesize{}FPN} & {\footnotesize{}ResNet101} & {\footnotesize{}1333x800} & {\footnotesize{}72.0} & {\footnotesize{}39.1}\tabularnewline
{\footnotesize{}6} & {\footnotesize{}COCO} & {\footnotesize{}Cascade-RCNN} & {\footnotesize{}ResNet50} & {\footnotesize{}800x600} & {\footnotesize{}54.9} & {\footnotesize{}39.3}\tabularnewline
\hline 
\end{tabular}{\footnotesize\par}
\par\end{centering}
\vspace{1mm}

\caption{Preliminary empirical experiments. Inference time is tested on one
V100 GPU. The performance of a detection model is highly related to
the dataset (Exp1-4). Better combination of modules and input resolution
can leads to an efficient detection system (Exp 5\&6).}

\vspace{-2mm}
\end{table}

1) One-stage detector is not always faster than two-stage detector.
Although RetinaNet \cite{he2016deep} is faster than FPN \cite{lin2017feature}
on VOC (Exp 3\&4), it is slower and worse than FPN on COCO (Exp 1\&2).

2) Reasonable combination of modules and input resolution can lead
to an efficient detection system. Generally, Cascade-RCNN is slower
than FPN with the same backbone since it has 2 more cascade heads.
However, with a better combination of modules and input resolution,
CascadeRCNN with ResNet50 can be faster and more accurate than FPN
with ResNet101 (Exp 5 \& 6).

It can be found that customizing different modules and input-size
is crucial for real-time object detection system for task specific
datasets. Thus we present the SM-NAS for searching an efficient combination
of modules and better modular-level architecture for object detection.

\subsection{NAS Pipeline}

As in Figure \ref{fig:framework-graph}, we propose a coarse-to-fine
searching pipeline: 1) Structural-level searching stage first aims
to find an efficient combination of different modules; 2) Modular-level
search stage then evolves each specific module and push forward to
a faster task-specific network. Moreover, we explore a strategy of
fast training from scratch for the detection task, which can directly
search a detection backbone without pre-trained models or any proxy
task.

\subsubsection{Stage-one: Structural-level Searching}

Modern object detection systems can be decoupled into four components:
backbone, feature fusion neck, region proposal network (RPN), and
RCNN head. We consider putting different popular and latest choices
of modules into the search space to cover many popular designs.

\textbf{Backbone.} Commonly used backbones are included in the search
space: ResNet \cite{he2016deep} (ResNet18, ResNet34, ResNet50 and
ResNet101), ResNeXt \cite{xie2017aggregated} (ResNeXt50, ResNeXt101)
and MobileNet V2 \cite{sandler2018mobilenetv2}. During Stage-one,
we loaded the backbones pre-trained from ImageNet \cite{russakovsky2015imagenet}
for fast convergence.

\textbf{Feature fusion neck. }Features from different layers are commonly
used to predict objects across various sizes. The feature fusion neck
aims at conducting feature fusion for better prediction. Here, we
use \{$P_{1}$, $P_{2}$, $P_{3}$, $P_{4}$\} to denote feature levels
generated by the backbone e.g. ResNet. From $P_{1}$ to $P_{4}$,
the spatial size is gradually down-sampled with factor 2. We further
add two smaller $P_{5}$ and $P_{6}$ feature maps downsampled from
$P_{4}$ following RetinaNet \cite{lin2018focal}. The search space
contains: no FPN (the original Faster RCNN setting) and FPN with different
choices of input and output feature levels (ranging from $P_{1}$
to $P_{6}$).

\textbf{Region proposal network (RPN). }RPN generates multiple foreground
proposals within each feature map and only exists in two-stage detectors.
Our search space is chosen to be: no RPN (one-stage detectors); with
RPN; with Guided anchoring RPN \cite{wang2019region}.

RPN generates multiple foreground anchor proposals within each feature
map and only exists in two-stage detectors. Our search space is chosen
to be: no RPN (one-stage detectors); with RPN; with Guided anchoring
RPN \cite{wang2019region}.

\textbf{RCNN head. }RCNN head refines the objects location and predicts
final classification results. \cite{cai2017cascade} proposed cascade
RCNN heads to iterative refine the detection results, which has been
proved to be useful yet requiring more computational resources. Thus,
we consider regular RCNN head \cite{Ren2015,Lin2017a}, RetinaNet
head \cite{lin2017focal}, and cascade RCNN heads with different number
of heads (2 to 4) as our search space to exam the accuracy/speed trade-off.
Note that our search space covers both one-stage and two-stage detection
systems.

\textbf{Input Resolution.}Furthermore, the input resolution is closely
related to the accuracy and speed. \cite{qin2019thundernet} also
suggested that input resolution should match the capability of the
backbone, which is not measurable in practice. Intuitively, we thus
add input resolution in our search space to find the best matching
with different models: 512x512, 800x600, 1080x720 and 1333x800.

Inference time is then evaluated for each combination of modules.
Together with the accuracy on validation dataset, a Pareto front is
then generated showing the optimal structures of the detector under
different resource constraints.

\subsubsection{Stage-two: Modular-level Search\label{subsec:Stage-two:-Modular-level-Search}}

On the Pareto front generated by Stage-one, we can pick up several
efficient detection structures with different combination of modules.
Then in Stage-two, we search the detailed architecture for each module
and lift the boundary of speed/accuracy tradeoff of the selected structures.

\cite{qin2019thundernet} suggested that in detection backbone, early-stage
feature maps are larger with low-level features which describe spatial
details, while late-stage feature maps are smaller with high-level
features which are more discriminative. Localization subtask is sensitive
to low-level features while high-level features are crucial for classification.
Thus, a natural question is to ask how to leverage the computational
cost over different stages to obtain an optimal design for detection.
Therefore, inside the backbone, we design a flexible search space
to find the optimal base channel size, as well as the position of
down-sampling and channel-raising.

As shown in Figure \ref{fig:framework-graph}, the Stage-two backbone
search space consists of 5 stages, each of which refers to a bench
of convolutional blocks fed by the features with the same resolution.
The spatial size of stage 1 to 5 is gradually downsampled with factor
2. As suggested in \cite{li2019partial}, we fix stage 1 and the first
layer of stage 2 to be a 3x3 conv (stride=2). We use the same block
setting (basic/bottleneck residual block, ResNeXt block or MBblock
\cite{sandler2018mobilenetv2}) as the structures selected from the
result of Stage-one. For example, if the candidate model selected
from Stage-one\textquoteright s Pareto front is with ResNet101 as
the backbone, we will use the corresponding bottleneck residual block
as its search space.

Furthermore, the backbone architecture encoding string is like \textquotedblleft basicblock
54 1211-211-1111-12111\textquotedblright{} where the first placeholder
encodes the block setting; 54 is the base channel size; \textquotedblleft -\textquotedblright{}
separates each stage with different resolution; \textquotedblleft 1\textquotedblright{}
means regular block with no change of channels and \textquotedblleft 2\textquotedblright{}
indicated the number of base channels is doubled in this block. The
base channel size is chosen from ${48,56,64,72}$. Since there is
no pre-trained model available for customized backbones, we use a
fast train-from-scratch technique instead which will be elaborated
in the next section.

Besides the flexible backbone, we also adjust the channel size of
the FPN during the Stage-two search. The input channel size is chosen
from ${128,256,512}$ and the channels of the head is adjusted correspondingly.
Thus, the objective of Stage-two is to further refine the detailed
modular structure of the selected efficient architectures.

\subsection{Train from scratch and fast evaluate the architecture\label{subsec:Train-from-scratch}}

Most of the detection models require initialization of backbone from
the ImageNet \cite{russakovsky2015imagenet} pre-trained models during
training. Any modification on the structure of backbone requires training
again on the ImageNet, which makes it harder to evaluate the performance
of a customized backbone. This paradigm hinders the development of
efficient NAS for detection problem. \cite{shen2017dsod} first explores
the possibility of training a detector from scratch by the deeply
supervised networks and dense connections. \cite{he2018rethinking}
and ScratchDet \cite{zhu2018scratchdet} find that normalization play
an significant role in training from scratch and a longer training
can then help to catch up pre-trained counterparts. Inspired by those
works, we conjecture the difficulty from two factors and try to fix
them:

1) Inaccurate Batch Normalization because of smaller batch size: During
the training, the batch-size is usually very small because of high
GPU consumption, which leads to inaccurate estimation of the batch
statistics and increasing the model error dramatically \cite{wu2018group}.
To alleviate this problem, we use Group Normalization (GN) instead
of standard BN since GN is not sensitive to the batch size.

2) Complexity of the loss landscape: \cite{shen2017dsod} suggested
that the multiple loss and ROI pooling layer in detection hinder the
gradient of region-level backward to the backbone. Significant loss
jitter or gradient explosion are often observed during training from
scratch. BN has been proved to be an effective solution of the problem
through significantly smoothing the optimization landscape \cite{santurkar2018does,zhu2018scratchdet}.
Instead of using BN, which is not suitable for small batch size training,
we adopt Weight Standardization (WS) \cite{qiao2019weight} for the
weights in the convolution layers to further smooth the loss landscape.

Experiments in the later section show that with GN and WS, a much
larger learning rate can be adopted, thus enabling us to train a detection
network from scratch even faster than the pre-trained counterparts.

\begin{figure*}[t]
\begin{centering}
\includegraphics[scale=0.38]{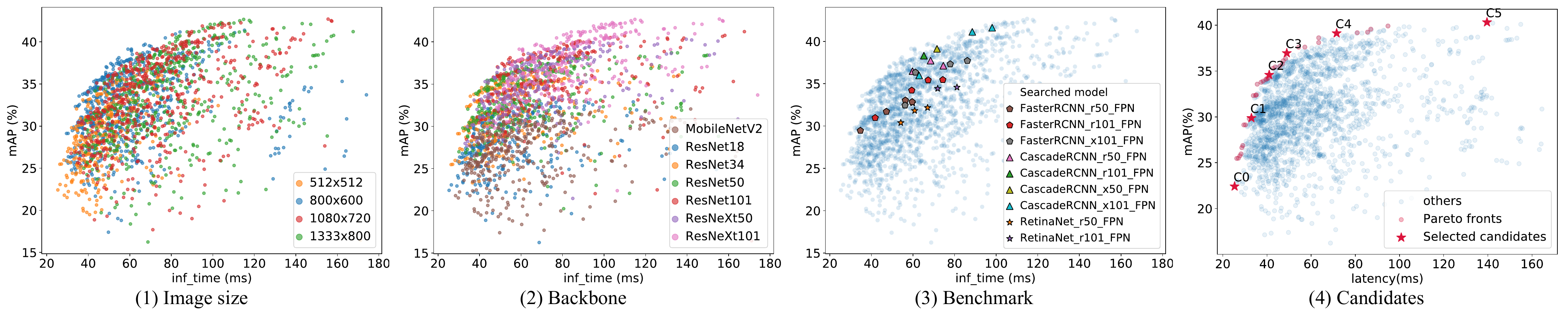}
\par\end{centering}
\vspace{-2mm}

\caption{\label{fig:stage1} Intermediate results for Stage-one: Structural-level
Searching. Comparison of mAP and inference time of all the architectures
searched on COCO. Inference time is tested on one V100 GPU. It can
be found our searching already found many structures dominate state-of-the-art
objectors. On the Pareto front, we pick 6 models (C0 to C5) and further
search for better modular-level architectures in Stage-two.}

\vspace{-2mm}
\end{figure*}

\subsection{Multi-objective Searching Algorithm}

For each stage, we aims at generating a Pareto front showing the optimal
trade-off between accuracy and different computation constrains. To
generate the Pareto front, we use nondominate sorting to determinate
whether one model dominates another in terms of both efficiency and
accuracy. In Stage-one, we use inference time on one V100 GPU as the
efficiency metric to roughly compare the actual performance between
different structures. In Stage-two, we use FLOPs instead of actual
time since FLOPs is more accurate than inference time to compare different
backbones with the same kind of block (the inference time has some
variation because of the GPU condition). Moreover, FLOPs is able to
keep the consistency of rank when changing the BN to GN+WS during
searching in Stage-two.

The architecture search step is based on: 1) the evolutionary algorithm
to mutate the best architecture on the Pareto front; 2) Partial Order
Pruning method \cite{li2019partial} to prune the architecture search
space with the prior knowledge that deeper models and wider models
are better. Our algorithm can be parallelized on multiple computation
nodes (each has 8 V100 GPUs) and lift the Pareto front simultaneously.

\section{Experiments}

\begin{figure}
\begin{centering}
\includegraphics[scale=0.28]{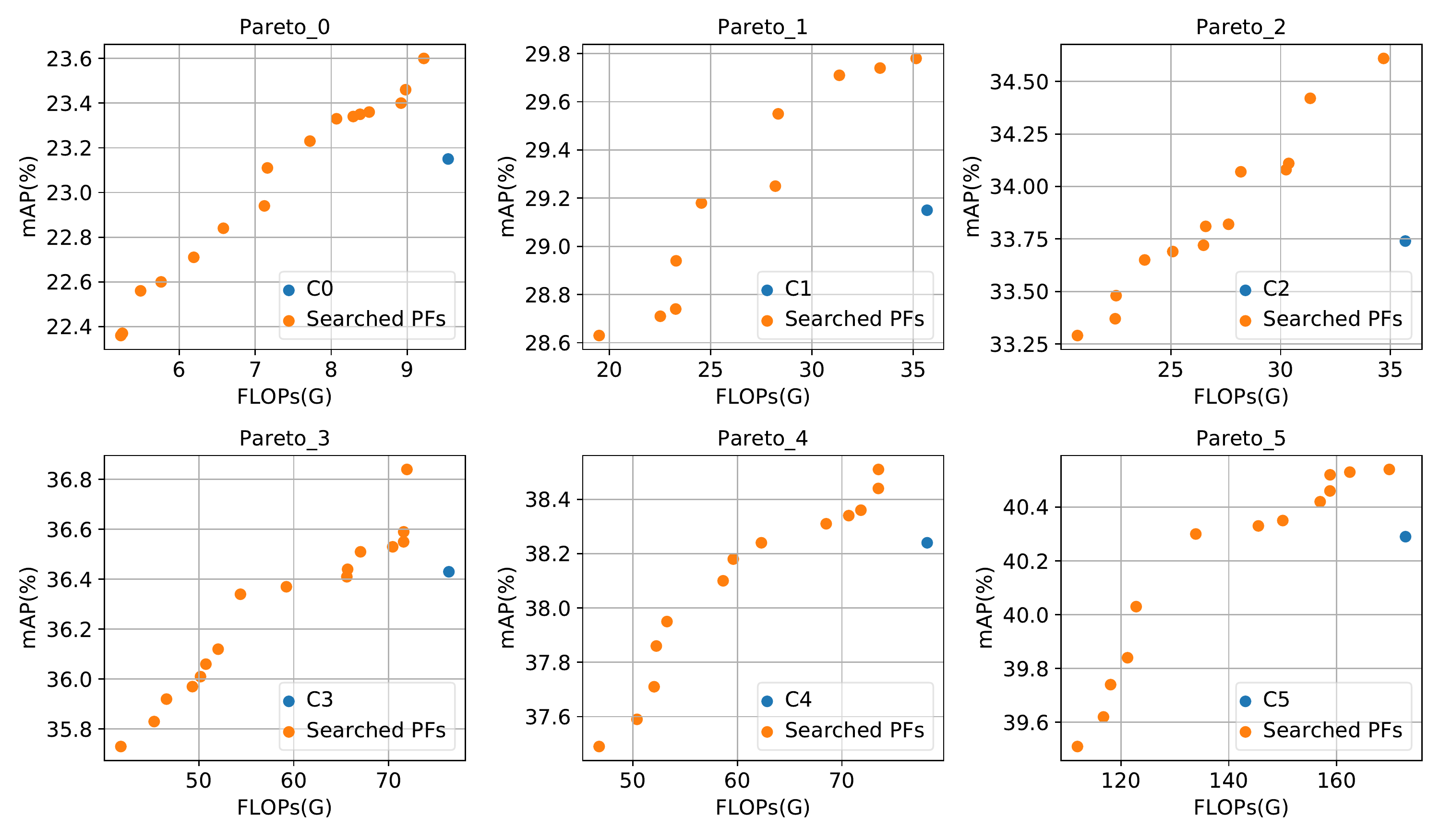}
\par\end{centering}
\vspace{-2mm}

\caption{\label{fig:stage2} Intermediate results for Modular-level Search.
The architectures with blue dot are the selected model C0-C5 based
on the previous Stage-one. The orange dots are architectures forming
the Pareto front found by our algorithm.}

\vspace{-1mm}
\end{figure}

\begin{table*}
\begin{centering}
\tabcolsep 0.02in{\scriptsize{}}%
\begin{tabular}{c|c|c|c|c|c|c|c|c}
\hline 
{\scriptsize{}Model} & {\scriptsize{}Input size} & {\scriptsize{}Backbone} & {\scriptsize{}Neck} & {\scriptsize{}RPN} & {\scriptsize{}RCNN Head} & {\scriptsize{}Backbone FLOPs} & {\scriptsize{}Time (ms)} & {\scriptsize{}mAP}\tabularnewline
\hline 
{\scriptsize{}E0} & {\scriptsize{}512x512} & {\scriptsize{}basicblock\_64\_1-21-21-12} & {\scriptsize{}FPN($P_{2}$-$P_{5}$, c=128)} & {\scriptsize{}RPN} & {\scriptsize{}2FC} & {\scriptsize{}7.2G (0.75)} & {\scriptsize{}24.5} & {\scriptsize{}27.1}\tabularnewline
{\scriptsize{}E1} & {\scriptsize{}800x600} & {\scriptsize{}basicblock\_56\_111-2111-2-111112} & {\scriptsize{}FPN($P_{2}$-$P_{5}$, c=256)} & {\scriptsize{}RPN} & {\scriptsize{}2FC} & {\scriptsize{}28.3G (0.79)} & {\scriptsize{}32.2} & {\scriptsize{}34.3}\tabularnewline
{\scriptsize{}E2} & {\scriptsize{}800x600} & {\scriptsize{}basicblock\_48\_12-11111-211-1112} & {\scriptsize{}FPN($P_{1}$-$P_{5}$, c=128)} & {\scriptsize{}RPN} & {\scriptsize{}Cascade(n=3)} & {\scriptsize{}23.8G (0.67)} & {\scriptsize{}39.5} & {\scriptsize{}40.1}\tabularnewline
{\scriptsize{}E3} & {\scriptsize{}800x600} & {\scriptsize{}bottleneck\_56\_211-111111111-2111111-11112111} & {\scriptsize{}FPN($P_{1}$-$P_{5}$, c=128)} & {\scriptsize{}RPN} & {\scriptsize{}Cascade(n=3)} & {\scriptsize{}59.2G (0.78)} & {\scriptsize{}50.7} & {\scriptsize{}42.7}\tabularnewline
{\scriptsize{}E4} & {\scriptsize{}800x600} & {\scriptsize{}Xbottleneck\_56\_21-21-111111111111111-2111111} & {\scriptsize{}FPN($P_{1}$-$P_{5}$, c=256)} & {\scriptsize{}GA-RPN} & {\scriptsize{}Cascade(n=3)} & {\scriptsize{}73.5G (0.96)} & {\scriptsize{}80.2} & {\scriptsize{}43.9}\tabularnewline
{\scriptsize{}E5} & {\scriptsize{}1333x800} & {\scriptsize{}Xbottleneck\_56\_21-21-11111111111111-21111111} & {\scriptsize{}FPN($P_{1}$-$P_{5}$, c=256)} & {\scriptsize{}GA-RPN} & {\scriptsize{}Cascade(n=3)} & {\scriptsize{}162.45G (0.94)} & {\scriptsize{}108.1} & {\scriptsize{}46.1}\tabularnewline
\hline 
\end{tabular}{\scriptsize\par}
\par\end{centering}
\vspace{1mm}

\caption{\label{tab:The-final-searched}Detailed architecture of the final
SM-NAS models from E0 to E5. For the backbone, basicblock and bottleneck
follow the same as in ResNet \cite{He2016} and Xbottleneck refers
to the block setting of ResNeXt \cite{Xie2017}. For Neck, $P_{2}$-$P_{5}$
and \textquotedblleft c\textquotedblright{} denotes the choice and
the channels of output feature levels in FPN. For RCNN head, \textquotedblleft 2FC\textquotedblright{}
is the regular setting of two shared fully connected layer; \textquotedblleft n\textquotedblright{}
means the stages of the cascade head.}
\vspace{-2mm}
\end{table*}

\subsection{Architecture Search Implementation Details and intermediate results}

We conduct architecture search on the well-known \textbf{COCO} \cite{lin2014microsoft}
dataset, which contains 80 object classes with 118K images for training,
5K for evaluation. For Stage-one, we consider a totally $1.1\times10^{4}$
combination of modules. For Stage-two, the search space is much larger,
containing about $5.0\times10^{12}$ unique paths. We conduct all
experiments using Pytorch \cite{paszke2017automatic,mmdetection2018},
multiple computational nodes with 8 V100 cards on each server. To
measure the inference speed, we run all the testing images on one
V100 GPU and take the average inference time for comparison. All experiments
are performed under CUDA 9.0 and CUDNN 7.0.

\subsubsection{Implementation Details for Stage-one.}

During searching, we first generate some initial models with a random
combination of modules. Then evolutionary algorithm is used to mutate
the best architecture on the Pareto front and provides candidate models.
During architectures evaluation, we use SGD optimizer with cosine
decay learning rate from 0.04 to 0.0001, momentum 0.9 and $10^{-4}$
as weight decay. Pre-trained models on ImageNet \cite{russakovsky2015imagenet}
are used as our backbone for fast convergence. Empirically, we found
that training with 5 epochs can separate good models from bad models.
In this stage, we evaluate about 500 architectures and it takes about
2000 GPU hours for the whole searching process.

\textbf{Intermediate results for Stage-one. }The first two figures
in \ref{fig:stage1} show the comparison of mAP and inference time
of the architectures searched on COCO. From Figure \ref{fig:stage1}-1,
it can be found that different input resolution can variate the speed
and accuracy. We also found that MobileNet V2 is dominated by other
models although it has mush less FLOPs in Figure \ref{fig:stage1}-2.
This is because it has higher memory access cost thus is slower in
practice \cite{li2019partial}. Therefore, using the direct metric,
i.e. inference time, rather than approximate metric such as FLOPs
is necessary for achieving the best speed/accuracy trade-off and our
searching found some structures dominate classic detectors. From Figure
\ref{fig:stage1}-3, it can be found that our searching already found
some structures dominate classic objectors. On the generated Pareto
front, we pick 6 models (C0 to C5) and further search for the better
modular-level architectures in Stage-two.

\subsubsection{Implementation Details for Stage-two.}

During Stage-two, we use the training strategy with GN and WS methods
discussed in the previous section. We use cosine decay learning rate
ranging from 0.24 to 0.0001 with batch size 8 on each GPU. The model
is trained with 9 epochs to fully explore the different modular-level
structures. It is worth mention that we directly search on the COCO
without pre-trained models. In Stage-two, we evaluate about 300 architectures
for each group and use about 2500 GPU hours.

\textbf{Intermediate results for Stage-two.} Figure \ref{fig:stage2}
shows mAP/speed improvement of the searched models compared to the
optimal model selected in Stage-one. It can be found that SM-NAS can
further push the Pareto front to a better trade-off of speed/accuracy.

\begin{table*}[t]
\vspace{-1mm}

\begin{centering}
{\scriptsize{}}%
\begin{tabular}{c|c|c|c|c|c|c|c|c|c}
\hline 
{\scriptsize{}Method} & {\scriptsize{}Backbone} & {\scriptsize{}Input size} & {\scriptsize{}Inf time (ms)} & {\scriptsize{}$\mathrm{\mathrm{\textrm{AP}}}$} & {\scriptsize{}AP$_{50}$} & {\scriptsize{}AP$_{75}$} & {\scriptsize{}AP$_{S}$} & {\scriptsize{}AP$_{M}$} & {\scriptsize{}AP$_{L}$}\tabularnewline
\hline 
{\scriptsize{}YOLO v3\cite{redmon2018yolov3}} & {\scriptsize{}DarkNet-53} & {\scriptsize{}608x608} & {\scriptsize{}51.0 (TitanX)} & {\scriptsize{}33.0} & {\scriptsize{}57.9} & {\scriptsize{}34.4} & {\scriptsize{}18.3} & {\scriptsize{}35.4} & {\scriptsize{}41.9}\tabularnewline
{\scriptsize{}DSSD513\cite{fu2017dssd}} & {\scriptsize{}ResNet101} & {\scriptsize{}513x513} & {\scriptsize{}-} & {\scriptsize{}33.2} & {\scriptsize{}53.3} & {\scriptsize{}35.2} & {\scriptsize{}13.0} & {\scriptsize{}35.4} & {\scriptsize{}51.1}\tabularnewline
{\scriptsize{}RetinaNet\cite{lin2018focal}} & {\scriptsize{}ResNet101-FPN} & {\scriptsize{}1333x800} & {\scriptsize{}91.7 (V100)} & {\scriptsize{}39.1} & {\scriptsize{}59.1} & {\scriptsize{}42.3} & {\scriptsize{}21.7} & {\scriptsize{}42.7} & {\scriptsize{}50.2}\tabularnewline
{\scriptsize{}FSAF\cite{zhu2019feature}} & {\scriptsize{}ResNet101-FPN} & {\scriptsize{}1333x800} & {\scriptsize{}92.5 (V100)} & {\scriptsize{}40.9} & {\scriptsize{}61.5} & {\scriptsize{}44.0} & {\scriptsize{}24.0} & {\scriptsize{}44.2} & {\scriptsize{}51.3}\tabularnewline
{\scriptsize{}CornerNet\cite{law2018cornernet}} & {\scriptsize{}Hourglass-104} & {\scriptsize{}512x512} & {\scriptsize{}244.0 (TitanX)} & {\scriptsize{}40.5} & {\scriptsize{}56.5} & {\scriptsize{}43.1} & {\scriptsize{}19.4} & {\scriptsize{}42.7} & {\scriptsize{}53.9}\tabularnewline
{\scriptsize{}CenterNet\cite{zhou2019objects}} & {\scriptsize{}Hourglass-104} & {\scriptsize{}512x512} & {\scriptsize{}126.0 (V100)} & {\scriptsize{}42.1} & {\scriptsize{}61.1} & {\scriptsize{}45.9} & {\scriptsize{}24.1} & {\scriptsize{}45.5} & {\scriptsize{}52.8}\tabularnewline
{\scriptsize{}AlignDet\cite{chen2019revisiting}} & {\scriptsize{}ResNet101-FPN} & {\scriptsize{}1333x800} & {\scriptsize{}110.0 (P100)} & {\scriptsize{}42.0} & {\scriptsize{}62.4} & {\scriptsize{}46.5} & {\scriptsize{}24.6} & {\scriptsize{}44.8} & {\scriptsize{}53.3}\tabularnewline
\hline 
{\scriptsize{}GA-Faster RCNN\cite{wang2019region}} & {\scriptsize{}ResNet50-FPN} & {\scriptsize{}1333x800} & {\scriptsize{}104.2 (V100)} & {\scriptsize{}39.8} & {\scriptsize{}59.2} & {\scriptsize{}43.5} & {\scriptsize{}21.8} & {\scriptsize{}42.6} & {\scriptsize{}50.7}\tabularnewline
{\scriptsize{}Faster-RCNN\cite{ren2015faster}} & {\scriptsize{}ResNet101-FPN} & {\scriptsize{}1333x800} & {\scriptsize{}84.0 (V100)} & {\scriptsize{}39.4} & {\scriptsize{}-} & {\scriptsize{}-} & {\scriptsize{}-} & {\scriptsize{}-} & {\scriptsize{}-}\tabularnewline
{\scriptsize{}Mask-RCNN\cite{he2017mask}} & {\scriptsize{}ResNet101-FPN} & {\scriptsize{}1333x800} & {\scriptsize{}105.0 (V100)} & {\scriptsize{}40.2} & {\scriptsize{}-} & {\scriptsize{}-} & {\scriptsize{}-} & {\scriptsize{}-} & {\scriptsize{}-}\tabularnewline
{\scriptsize{}Cascade-RCNN\cite{cai2017cascade}} & {\scriptsize{}ResNet101-FPN} & {\scriptsize{}1333x800} & {\scriptsize{}97.9 (V100)} & {\scriptsize{}42.8} & {\scriptsize{}62.1} & {\scriptsize{}46.3} & {\scriptsize{}23.7} & {\scriptsize{}45.5} & {\scriptsize{}55.2}\tabularnewline
{\scriptsize{}TridentNet\cite{li2019scale}} & {\scriptsize{}ResNet101} & {\scriptsize{}1333x800} & {\scriptsize{}588 (V100)} & {\scriptsize{}42.7} & {\scriptsize{}63.6} & {\scriptsize{}46.5} & {\scriptsize{}23.9} & {\scriptsize{}46.4} & {\scriptsize{}55.6}\tabularnewline
{\scriptsize{}TridentNet\cite{li2019scale}} & {\scriptsize{}ResNet101-deformable-FPN} & {\scriptsize{}1333x800} & {\scriptsize{}2498.3 (V100)} & {\scriptsize{}48.4} & {\scriptsize{}69.7} & {\scriptsize{}53.5} & {\scriptsize{}31.8} & {\scriptsize{}51.3} & {\scriptsize{}60.3}\tabularnewline
\hline 
{\scriptsize{}DetNAS\cite{chen2019detnas}} & {\scriptsize{}Searched Backbone} & {\scriptsize{}1333x800} & {\scriptsize{}-} & {\scriptsize{}42.0} & {\scriptsize{}63.9} & {\scriptsize{}45.8} & {\scriptsize{}24.9} & {\scriptsize{}45.1} & {\scriptsize{}56.8}\tabularnewline
{\scriptsize{}NAS-FPN\cite{Ghiasi_2019_CVPR}} & {\scriptsize{}ResNet50-FPN(@384)} & {\scriptsize{}1280x1280} & {\scriptsize{}198.7 (V100)} & {\scriptsize{}45.4} & {\scriptsize{}-} & {\scriptsize{}-} & {\scriptsize{}-} & {\scriptsize{}-} & {\scriptsize{}-}\tabularnewline
\hline 
{\scriptsize{}SM-NAS: E2} & {\scriptsize{}Searched Backbone} & {\scriptsize{}800x600} & \textbf{\scriptsize{}39.5}{\scriptsize{}(V100)} & \textbf{\scriptsize{}40.0} & {\scriptsize{}58.2} & {\scriptsize{}43.4} & {\scriptsize{}21.1} & {\scriptsize{}42.4} & {\scriptsize{}51.7}\tabularnewline
{\scriptsize{}SM-NAS: E3} & {\scriptsize{}Searched Backbone} & {\scriptsize{}800x600} & \textbf{\scriptsize{}50.7}{\scriptsize{}(V100)} & \textbf{\scriptsize{}42.8} & {\scriptsize{}61.2} & {\scriptsize{}46.5} & {\scriptsize{}23.5} & {\scriptsize{}45.5} & {\scriptsize{}55.6}\tabularnewline
{\scriptsize{}SM-NAS: E5} & {\scriptsize{}Searched Backbone} & {\scriptsize{}1333x800} & \textbf{\scriptsize{}108.1}{\scriptsize{}(V100)} & \textbf{\scriptsize{}45.9} & {\scriptsize{}64.6} & {\scriptsize{}49.6} & {\scriptsize{}27.1} & {\scriptsize{}49.0} & {\scriptsize{}58.0}\tabularnewline
\hline 
\end{tabular}{\scriptsize\par}
\par\end{centering}
\vspace{1mm}

\caption{\label{tab:state-of-art-coco}Comparison of mAP of the state-of-the-art
single-model on COCO test-dev. Our searched models dominate most SOTA
models in terms of speed/accuracy by a large margin.}

\vspace{-1mm}
\end{table*}

\subsection{Object Detection Results\label{subsec:Auto-neck-for-SSD}}

On the COCO dataset, the optimal architectures E0 to E5 are identified
with our two-stages search. We change the backbone back to BN and
no Weight Standardization mode since these practices will slow down
the inference time. We first pre-train those searched backbones on
ImageNet following common practice \cite{he2016deep} for fair comparison
with other methods. Then stochastic gradient descent (SGD) is performed
to train the full model on 8 GPUs with 4 images on each GPU. Following
the setting of 2x schedule \cite{he2018rethinking} the initial learning
rate is 0.04 (with a linear warm-up), and reduces two times ($\times0.1$)
during fine-tuning; $10^{-4}$ as weight decay; $0.9$ as momentum.
The training and testing is conducted with the searched optimal input
resolutions. Image flip and scale jitter is adopted for augmentation
during training, and evaluation procedure follows the COCO official
setting \cite{lin2014microsoft}.

\textbf{Detailed architectures of the final searched models}. Table
\ref{tab:The-final-searched} shows architecture details of the final
searched E0 to E5. Comparing the searched backbones with classical
ResNet/ResNeXt, we find that early stages in our models are very short
which is more efficient since feature maps in an early stage is very
large with a high computational cost. We also found that for high-performance
detectors E3-E5, raising channels usually happens in very early stage
which means that lower-level feature plays an important role for localization.The
classification performance of the backbone of E0 to E5 on ImageNet
can also be found in the supplementary materials. We can find the
searched backbones are also efficient in the classification task.

\textbf{Comparison with the state-of-the-art.} In Table \ref{tab:state-of-art-coco},
we make a detailed comparison with existing detectors: YOLOv3\cite{redmon2018yolov3},
DSSD\cite{fu2017dssd}, RetinaNet\cite{lin2018focal}, FSAF\cite{zhu2019feature},
CornerNet\cite{law2018cornernet}, CenterNet\cite{zhou2019objects},
AlignDet\cite{chen2019revisiting}, GA-FasterRCNN\cite{wang2019region},
Faster-RCNN\cite{ren2015faster}, Mask-RCNN\cite{he2017mask}, Cascade-RCNN\cite{cai2017cascade},
TridentNet\cite{li2019scale}, and NAS-FPN\cite{Ghiasi_2019_CVPR}.
Most reported results are tested with single V100 GPU (some models
marked with other GPU devices following the original papers). For
a fair comparison, multi-scale testing is not adopted for all methods.
From E0 to E5, SM-NAS constructs a Pareto front that dominates most
SOTA models as shown in Figure \ref{fig:benchmark-graph}. Our searched
models dominate most state-of-the-art models, demonstrating that SM-NAS
is able to find efficient real-time object detection systems.

\begin{table}
\vspace{-1mm}

\begin{centering}
\tabcolsep 0.02in{\scriptsize{}}%
\begin{tabular}{c|c|c|c|c|c|c}
\hline 
{\scriptsize{}id} & {\scriptsize{}Norm Method} & {\scriptsize{}ImageNet Pretrain} & {\scriptsize{}Epoch} & {\scriptsize{}Batchsize} & {\scriptsize{}lr} & {\scriptsize{}mAP}\tabularnewline
\hline 
{\scriptsize{}0} & {\scriptsize{}BN} & \textbf{\scriptsize{}$\checked$} & {\scriptsize{}12} & {\scriptsize{}2x8} & {\scriptsize{}0.02} & {\scriptsize{}36.5}\tabularnewline
{\scriptsize{}1} & {\scriptsize{}BN} & \textbf{\scriptsize{}$\checked$} & {\scriptsize{}24} & {\scriptsize{}2x8} & {\scriptsize{}0.02} & {\scriptsize{}37.4}\tabularnewline
{\scriptsize{}2} & {\scriptsize{}BN} & \textbf{\scriptsize{}$\times$} & {\scriptsize{}12} & {\scriptsize{}2x8} & {\scriptsize{}0.02} & {\scriptsize{}24.8}\tabularnewline
{\scriptsize{}3} & {\scriptsize{}BN} & \textbf{\scriptsize{}$\times$} & {\scriptsize{}12} & {\scriptsize{}8x8} & {\scriptsize{}0.20} & {\scriptsize{}28.3}\tabularnewline
{\scriptsize{}4} & {\scriptsize{}GN} & \textbf{\scriptsize{}$\times$} & {\scriptsize{}12} & {\scriptsize{}2x8} & {\scriptsize{}0.02} & {\scriptsize{}29.4}\tabularnewline
{\scriptsize{}5} & {\scriptsize{}GN+WS} & \textbf{\scriptsize{}$\times$} & {\scriptsize{}12} & {\scriptsize{}2x8} & {\scriptsize{}0.02} & {\scriptsize{}30.7}\tabularnewline
{\scriptsize{}6} & {\scriptsize{}GN+WS} & \textbf{\scriptsize{}$\times$} & {\scriptsize{}12} & {\scriptsize{}2x8} & {\scriptsize{}0.10} & {\scriptsize{}36.4}\tabularnewline
{\scriptsize{}7} & \textbf{\scriptsize{}GN+WS} & \textbf{\scriptsize{}$\times$} & \textbf{\scriptsize{}16} & \textbf{\scriptsize{}4x8} & \textbf{\scriptsize{}0.16} & \textbf{\scriptsize{}37.5}\tabularnewline
\hline 
\end{tabular}{\scriptsize\par}
\par\end{centering}
\vspace{1mm}

\caption{\label{tab:FPN-with-ResNet-50_train_from_scratch}FPN with ResNet-50
trained with different strategies, evaluated on COCO val. \textquotedblleft GN\textquotedblright{}
is group normalization by \cite{wu2018group}. \textquotedblleft WS\textquotedblright{}
is the Weight Standardization method by \cite{qiao2019weight}. We
found that with group normalization, Weight Standardization, larger
learning rate and batchsize, we can train a detection network from
scratch using less epochs than standard training procedure.}
\vspace{-1mm}
\end{table}

\textbf{Ablative study for strategies of training from scratch. }Since
in modular-level searching stage, we keep changing the backbone structure,
we need to find an optimal setting of training strategies for efficiently
training a detection network from scratch. Table \ref{tab:FPN-with-ResNet-50_train_from_scratch}
shows an ablative study of FPN with ResNet-50 trained with different
strategies, evaluated on COCO. Exp-0 and Exp-1 are the 1x and 2x standard
FPN training procedure following \cite{he2018rethinking}. Comparing
Exp-2\&3, with Exp-4, it can be found smaller batch size leads to
inaccurate batch normalization statistics. Using group normalization
can alleviate this problem and improve the mAP from 24.8 to 29.4.
From Exp-5, adding WS can further smooth the training and improve
mAP by 1.3. Furthermore, enlarging the learning rate and batch size
can increase the mAP to 37.5 in 16-epoch-training (see Exp 5\&6\&7).
Thus, we can train a detection network from scratch using fewer epochs
than the pre-trained counterparts.

\textbf{Architecture Transfer: VOC and BDD. }To evaluate the domain
transferability of the searched models, we transfer the searched architecture
E0-E3 from COCO to Pascal VOC and BDD. For \textbf{PASCAL VOC} dataset
\cite{Everingham10} with 20 object classes, training is performed
on the union of VOC 2007 trainval and VOC 2012 trainval (10K images)
and evaluation is on VOC 2007 test (4.9K images). We only report mAP
using IoU at 0.5. \textbf{Berkeley Deep Drive (BDD)} \cite{yu2018bdd100k}
is an autonomous driving dataset with 10 object classes, containing
about 70K images for training and 10K for evaluation. We use the same
training and testing configurations for a fare comparison. As shown
in Table \ref{tab:transfer}, on Pascal VOC, E0 reduces half of the
inference time compared to FPN with a higher mAP. For BDD, E3 is 17.4ms
faster than FPN. The searched architectures show good transferability.

\begin{table}
\vspace{-1mm}

\begin{centering}
\tabcolsep 0.015in{\scriptsize{}}%
\begin{tabular}{c|c|c|c|c}
\hline 
{\scriptsize{}Dataset} & {\scriptsize{}Input size} & {\scriptsize{}model} & {\scriptsize{}inf\_time (ms)} & {\scriptsize{}mAP}\tabularnewline
\hline 
\multirow{4}{*}{{\scriptsize{}VOC}} & \multirow{4}{*}{{\scriptsize{}800x600}} & {\scriptsize{}FPN w R50} & {\scriptsize{}38.4} & {\scriptsize{}80.4}\tabularnewline
 &  & {\scriptsize{}E0} & \textbf{\scriptsize{}21.5} & {\scriptsize{}81.4}\tabularnewline
 &  & {\scriptsize{}E1} & {\scriptsize{}35.9} & {\scriptsize{}83.7}\tabularnewline
 &  & {\scriptsize{}E3} & {\scriptsize{}47.0} & \textbf{\scriptsize{}84.4}\tabularnewline
\hline 
\multirow{4}{*}{{\scriptsize{}BDD}} & \multirow{4}{*}{{\scriptsize{}1333x800}} & {\scriptsize{}FPN w R101} & {\scriptsize{}84.6} & {\scriptsize{}36.9}\tabularnewline
 &  & {\scriptsize{}E0} & \textbf{\scriptsize{}27.8} & {\scriptsize{}30.2}\tabularnewline
 &  & {\scriptsize{}E1} & {\scriptsize{}45.2} & {\scriptsize{}37.9}\tabularnewline
 &  & {\scriptsize{}E3} & {\scriptsize{}67.2} & \textbf{\scriptsize{}39.6}\tabularnewline
\hline 
\end{tabular}{\scriptsize\par}
\par\end{centering}
\vspace{1mm}

\caption{\label{tab:transfer} Transferability of our models on PASCAL VOC
(VOC) and Berkeley Deep Drive dataset (BDD).}

\vspace{-1mm}
\end{table}

\begin{figure}
\vspace{-1mm}

\begin{centering}
\includegraphics[scale=0.45]{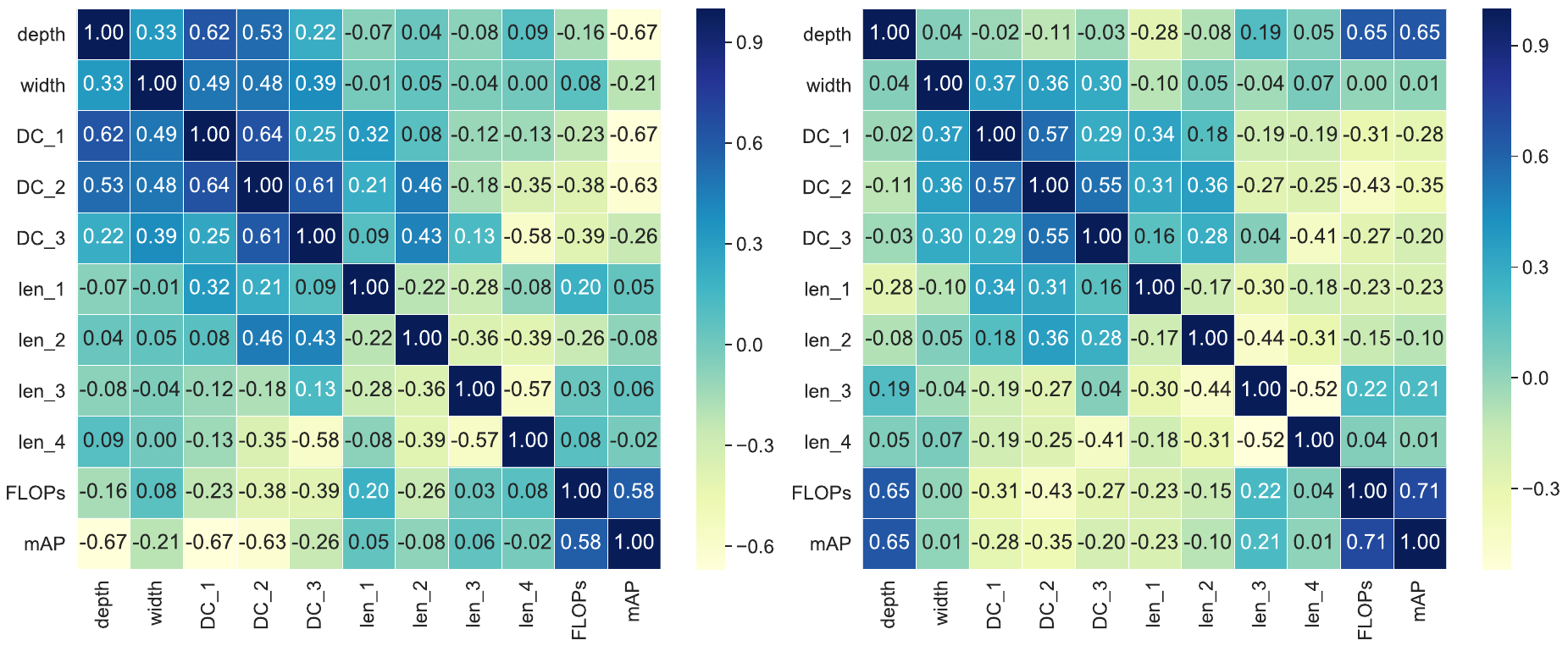}
\par\end{centering}
\vspace{-2mm}

\caption{\label{fig:correlation_all} Correlation between factors of the searched
models on COCO dataset. The left figure shows the results of Pareto
front 4 the right figure shows all the searched models. The depth
and width are the number of blocks and base channel size of backbone.
DC\_$x$ denotes the positions where the channel size is doubled;
and len\_$x$ denotes the proportion of the total blocks of the $x$th
stage.}
\end{figure}

\textbf{Correlation Analysis of the Architecture and mAP.} It is interesting
to analyze the correlation between the factors of backbone architecture
and mAP. Figure \ref{fig:correlation_all} shows the correlation between
factors of all the searched models on COCO dataset. The left figure
shows the results of Pareto front 4 in Stage-two. It can be found
that under the constraints of FLOPs, better architecture should decrease
the depth and put the computation budget in the low-level stage. The
right figure shows correlation for all the searched models. Depth
shows strong positive relation with mAP, raising channels in early
stage is good for detection. It is better to have a longer high-level
stage and shorter low-level stage.

\section{Conclusion}

We propose a detection NAS framework for searching both an efficient
combination of modules and better modular-level architectures for
object detection on a target device. The searched SM-NAS networks
achieve state-of-the-art speed/accuracy trade-off. The SM-NAS pipeline
can keep updating and adding new modules in the future.

{\small{}\bibliographystyle{ieee_fullname}
\bibliography{Know_network}
}{\small\par}

\section*{Supplementary}

\section*{Classification performance of our searched backbone of E0 to E5 on
ImageNet}

We further compare the Classification performance of our searched
backbone of E0 to E5 on ImageNet. We compare the FLOPS, memory access
cost (MAC) and total number of parameters with their counterparts
ResNet18, ResNet34, ResNet101 and ResNext101 in Table \ref{tab:The-final-searched-1}.
It can be found that all the searched backbone has a lower FLOPs,
MAC and total parameters with a higher Top-1 accuracy. More specifically,
the searched architecture nearly cut half of the FLOPs and total number
of parameters for E2, E3, E4 and E5. That\textquoteright s why it
is so efficient in the GPU. We can conclude that the searched architectures
are not only good at detection task, but also efficient on the classification.

\begin{figure*}
\begin{centering}
\tabcolsep 0.02in{\scriptsize{}}%
\begin{tabular}{c|c|c|ccc|c}
\hline 
{\small{}Model} & {\small{}Input size} & {\small{}Backbone} & {\small{}FLOPs(G)} & {\small{}MAC(M)} & {\small{}Parameters(M)} & {\small{}Top-1 Acc}\tabularnewline
\hline 
{\small{}R18} & {\small{}224x224} & {\small{}ResNet18} & {\small{}1.83} & {\small{}25.46} & {\small{}11.68} & {\small{}69.76}\tabularnewline
{\small{}E0} & {\small{}224x224} & {\small{}basic block\_64\_1-21-21-12} & \textbf{\small{}1.37} & \textbf{\small{}17.44} & \textbf{\small{}8.15} & \textbf{\small{}71.28}\tabularnewline
\hline 
{\small{}R34} & {\small{}224x224} & {\small{}ResNet34} & {\small{}3.68} & {\small{}41.85} & {\small{}21.80} & {\small{}73.30}\tabularnewline
{\small{}E1} & {\small{}224x224} & {\small{}basicblock\_56\_111-2111-2-111112} & {\small{}2.74} & {\small{}31.15} & {\small{}15.35} & \textbf{\small{}74.13}\tabularnewline
{\small{}E2} & {\small{}224x224} & {\small{}basicblock\_48\_12-11111-211-1112} & \textbf{\small{}2.46} & \textbf{\small{}26.28} & \textbf{\small{}9.92} & {\small{}73.66}\tabularnewline
\hline 
{\small{}R101} & {\small{}224x224} & {\small{}ResNet101} & {\small{}7.88} & {\small{}129.60} & {\small{}44.55} & {\small{}76.60}\tabularnewline
{\small{}E3} & {\small{}224x224} & {\small{}bottleneck\_56\_211-111111111-2111111-11112111} & \textbf{\small{}3.11} & \textbf{\small{}115.69} & \textbf{\small{}29.96} & {\small{}78.27}\tabularnewline
\hline 
{\small{}X101} & {\small{}224x224} & {\small{}ResNext101(32x4d)} & {\small{}16.55} & {\small{}233.75} & {\small{}88.79} & {\small{}78.80}\tabularnewline
{\small{}E4} & {\small{}224x224} & {\small{}Xbottleneck\_56\_21-21-111111111111111-2111111} & \textbf{\small{}7.58} & \textbf{\small{}135.75} & \textbf{\small{}43.14} & \textbf{\small{}79.07}\tabularnewline
{\small{}E5} & {\small{}224x224} & {\small{}Xbottleneck\_56\_21-21-11111111111111-21111111} & {\small{}7.58} & {\small{}137.48} & {\small{}45.74} & {\small{}79.03}\tabularnewline
\hline 
\end{tabular}{\scriptsize\par}
\par\end{centering}
\caption{\label{tab:The-final-searched-1} Classification performance of our
searched backbone of E0 to E5 on ImageNet. It can be found that all
the searched backbone has a lower FLOPs, MAC and total parameters
with a higher Top-1 accuracy.}
\vspace{-2mm}
\end{figure*}

\section*{More intermediate results for Stage-two}

In Stage-two, we conduct a further backbone search based on the module
combinations and input sizes searched in Stage-one. As a preliminary,
at the beginning of Stage-two we first train the candidate architectures
with vanilla backbones under the GN+WS setting to obtain baselines.
Table \ref{tab:more-stage2-result} shows the comparison between our
searched architectures (E0-E5) and the baselines. It can be found
that the searched backbones can considerably reduce FLOPs while keeping
a comparable mAP. Modular-search helps to further push the candidate
architectures to a better trade-off of speed/accuracy.

\begin{figure*}
\begin{centering}
\tabcolsep 0.02in{\scriptsize{}}%
\begin{tabular}{c|c|c|c|c}
\hline 
{\footnotesize{}Pareto id} & {\footnotesize{}Input size} & {\footnotesize{}Backbone} & {\footnotesize{}backbone FLOPs} & {\footnotesize{}mAP}\tabularnewline
\hline 
\multirow{2}{*}{{\footnotesize{}0}} & \multirow{2}{*}{{\footnotesize{}512x512}} & {\footnotesize{}ResNet18} & {\footnotesize{}9.54} & \textbf{\footnotesize{}23.15}\tabularnewline
 &  & {\footnotesize{}basicblock\_64\_1-21-21-12 (E0)} & \textbf{\footnotesize{}7.16} & {\footnotesize{}23.11}\tabularnewline
\hline 
\multirow{2}{*}{{\footnotesize{}1}} & \multirow{2}{*}{{\footnotesize{}800x600}} & {\footnotesize{}ResNet34} & {\footnotesize{}35.68} & {\footnotesize{}29.15}\tabularnewline
 &  & {\footnotesize{}basicblock\_56\_111-2111-2-111112 (E1)} & \textbf{\footnotesize{}28.33} & \textbf{\footnotesize{}29.55}\tabularnewline
\hline 
\multirow{2}{*}{{\footnotesize{}2}} & \multirow{2}{*}{{\footnotesize{}800x600}} & {\footnotesize{}ResNet34} & {\footnotesize{}35.68} & \textbf{\footnotesize{}33.74}\tabularnewline
 &  & {\footnotesize{}basicblock\_48\_12-11111-211-1112 (E2)} & \textbf{\footnotesize{}23.81} & {\footnotesize{}33.65}\tabularnewline
\hline 
\multirow{2}{*}{{\footnotesize{}3}} & \multirow{2}{*}{{\footnotesize{}800x600}} & {\footnotesize{}ResNet101} & {\footnotesize{}76.34} & {\footnotesize{}36.43}\tabularnewline
 &  & {\footnotesize{}bottleneck\_56\_211-111111111-2111111-11112111 (E3)} & \textbf{\footnotesize{}59.22} & \textbf{\footnotesize{}36.37}\tabularnewline
\hline 
\multirow{2}{*}{{\footnotesize{}4}} & \multirow{2}{*}{{\footnotesize{}800x600}} & {\footnotesize{}ResNeXt101} & {\footnotesize{}78.16} & {\footnotesize{}38.24}\tabularnewline
 &  & {\footnotesize{}Xbottleneck\_56\_21-21-111111111111111-2111111 (E4)} & \textbf{\footnotesize{}73.50} & \textbf{\footnotesize{}38.51}\tabularnewline
\hline 
\multirow{2}{*}{{\footnotesize{}5}} & \multirow{2}{*}{{\footnotesize{}1333x800}} & {\footnotesize{}ResNeXt101} & {\footnotesize{}172.78} & {\footnotesize{}40.29}\tabularnewline
 &  & {\footnotesize{}Xbottleneck\_56\_21-21-11111111111111-21111111 (E5)} & \textbf{\footnotesize{}162.45} & \textbf{\footnotesize{}40.53}\tabularnewline
\hline 
\end{tabular}{\scriptsize\par}
\par\end{centering}
\caption{\label{tab:more-stage2-result} More intermediate results for Stage-two.
The Pareto id from 0 to 5 refers to the search experiments based on
the corresponding candidates (described in Section4.1). ResNet and
ResNeXt represent the baseline results obtained by directly training
the Stage-one searched architectures under the Stage-two setting (GN+WS).}
 \vspace{-2mm}
\end{figure*}

\section*{More Correlation Results}

\begin{figure*}
\begin{centering}
\includegraphics[scale=0.35]{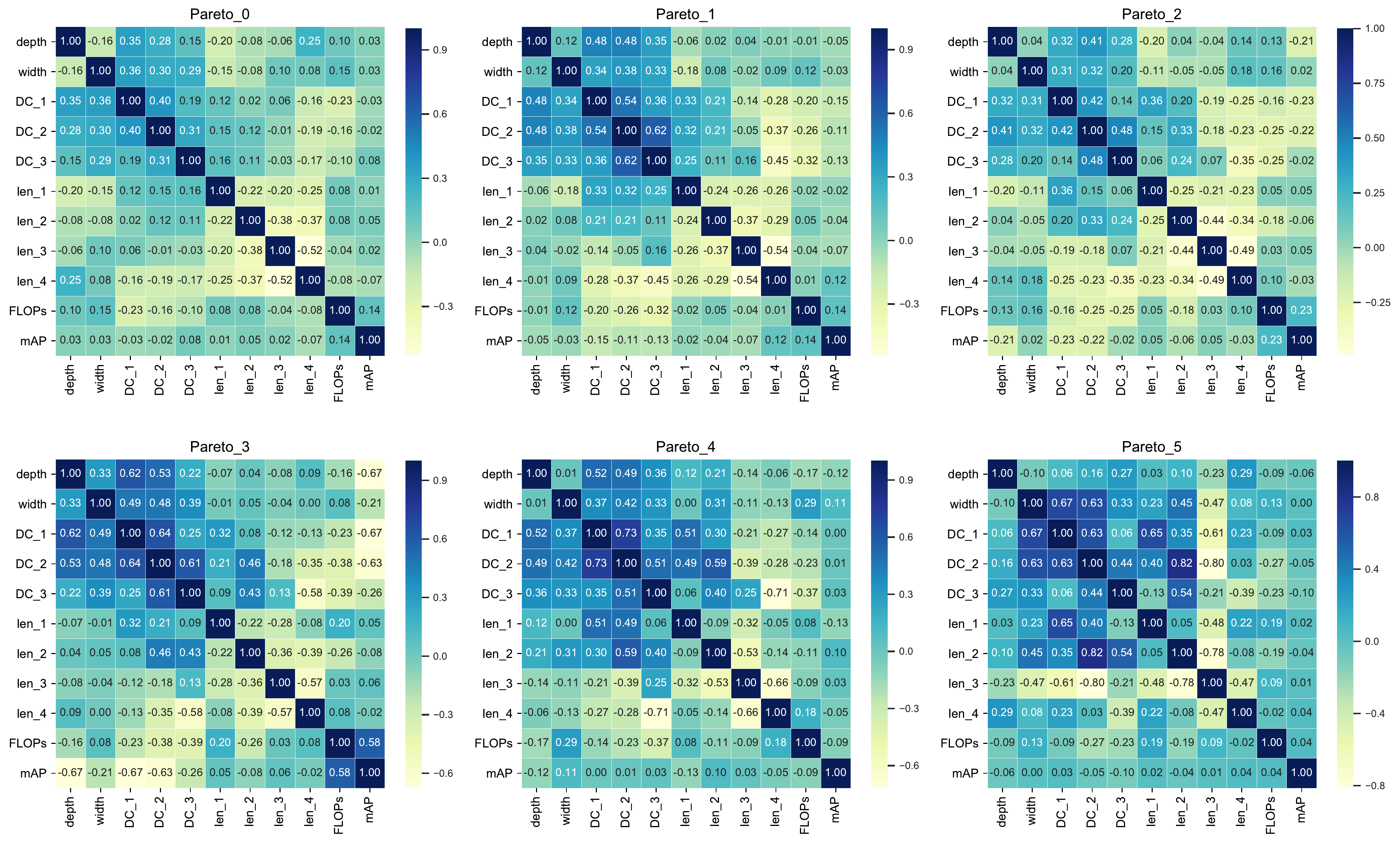}
\par\end{centering}
\caption{\label{fig:correlation}Correlation matrix of each Pareto fronts in
Stage-two. The depth and width are the number of blocks and base channel
size of backbone. DC\_$x$ denotes the positions which double the
channel size; and len\_$x$ denotes the proportion of the total blocks
of $x$th stage.}
\end{figure*}

Figure \ref{fig:correlation} shows the correlation coefficients between
factors of all the searched models on COCO dataset for all the Pareto
fronts in Stage-two. From Pareto front 0 to Pareto front 5, it can
be found that the correlation coefficients become more significant
which indicates the larger models tends to have specific patterns.
For small model, the mAP is positive correlated to the depth. However,
when the model becomes larger, the depth is negative related to the
mAP. It can be also found that under the constraints of FLOPs, better
architecture should decrease the depth and put the computation budget
in the low-level stage.

\section*{Qualitative Results and Comparison}

\begin{figure*}
\begin{centering}
\includegraphics[height=2.6cm]{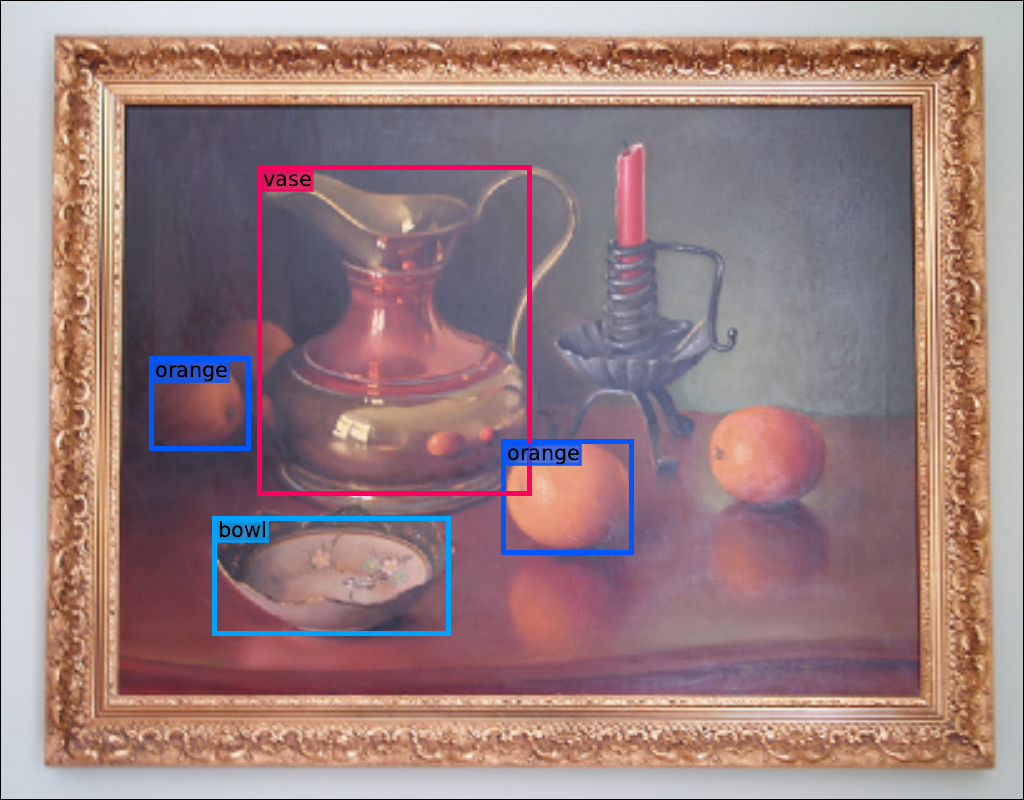}\includegraphics[height=2.6cm]{coco_fpn_0126}\includegraphics[height=2.6cm]{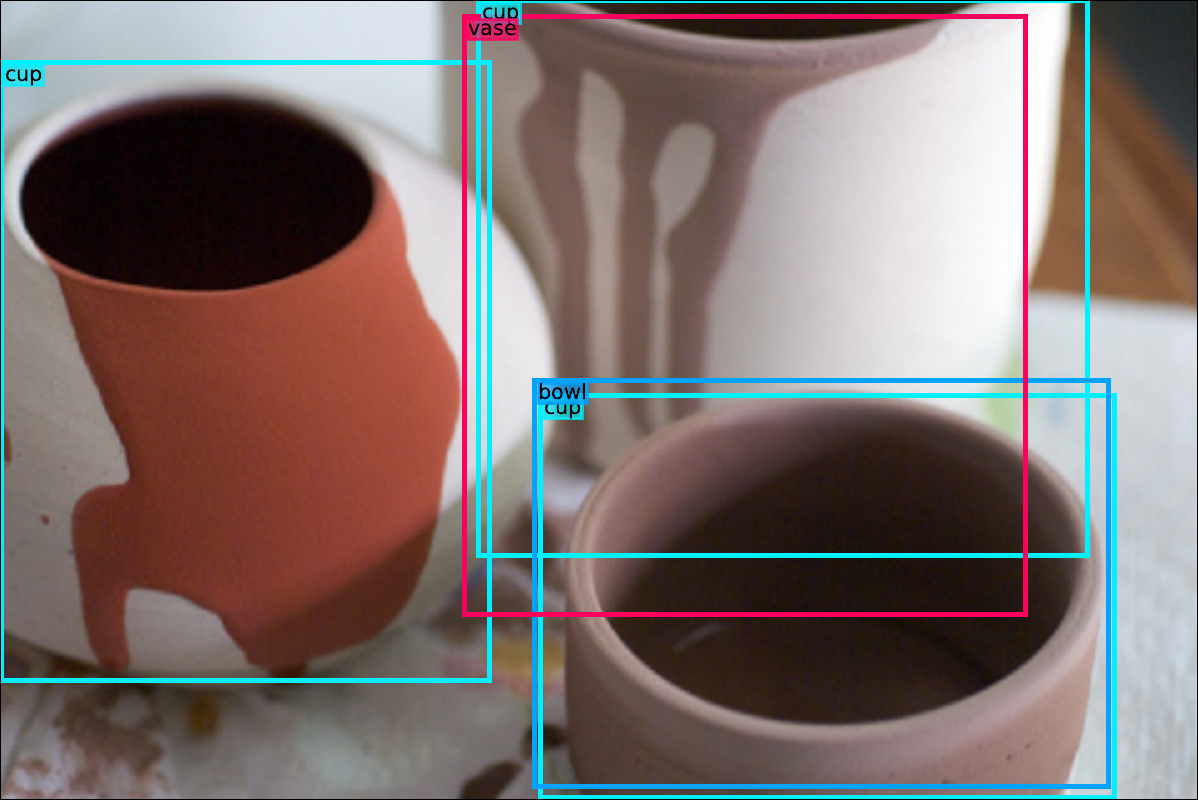}\includegraphics[height=2.6cm]{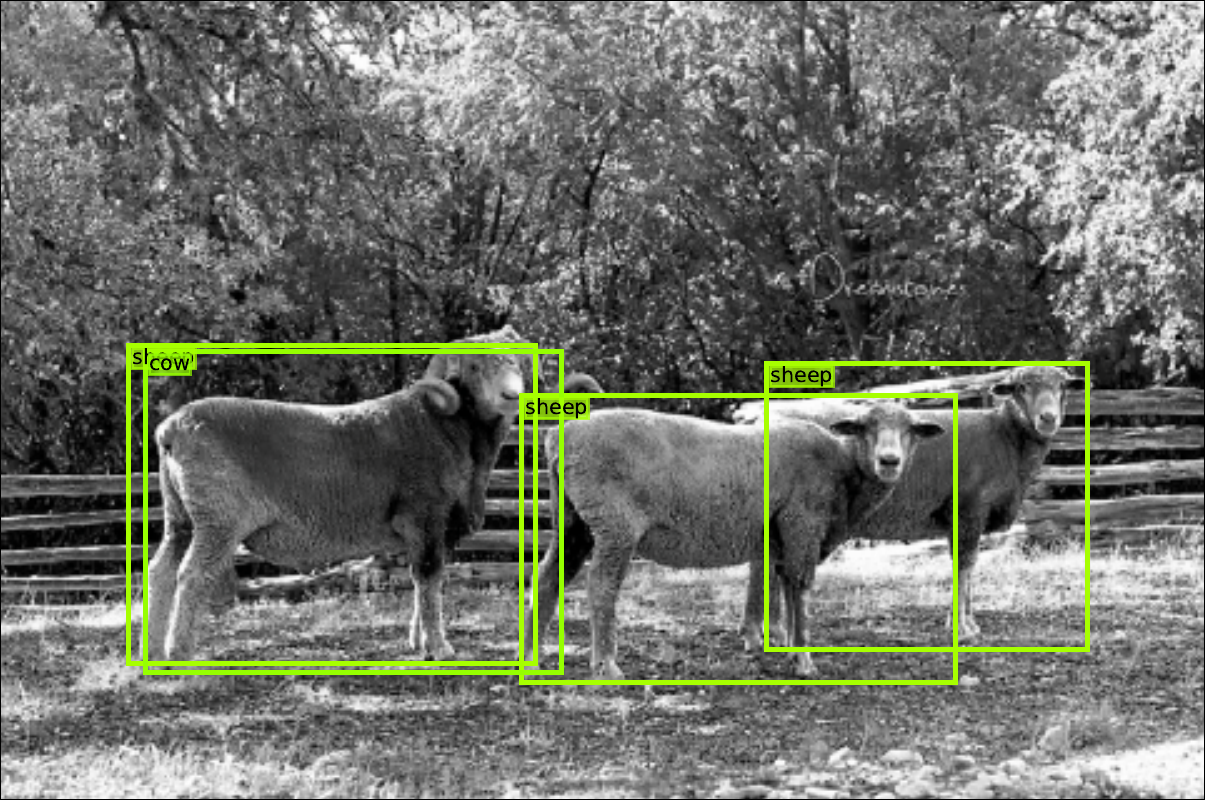}\includegraphics[height=2.6cm]{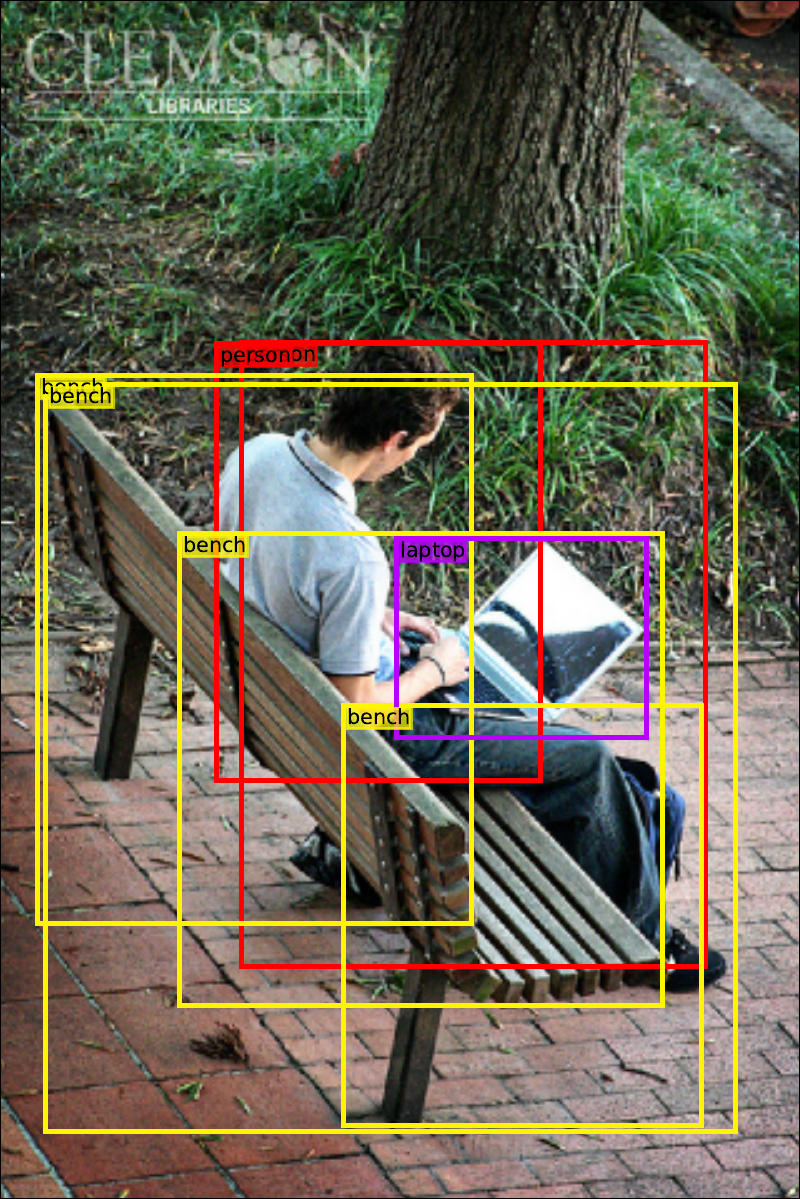}
\par\end{centering}
\begin{centering}
Vanilla FPN with ResNet101
\par\end{centering}
\vspace{1.5mm}

\begin{centering}
\includegraphics[height=2.6cm]{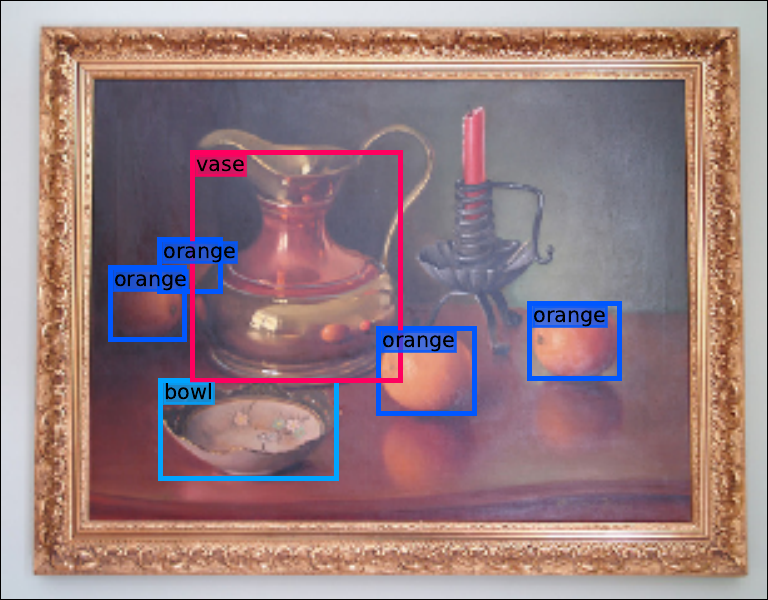}\includegraphics[height=2.6cm]{coco_e3_0126}\includegraphics[height=2.6cm]{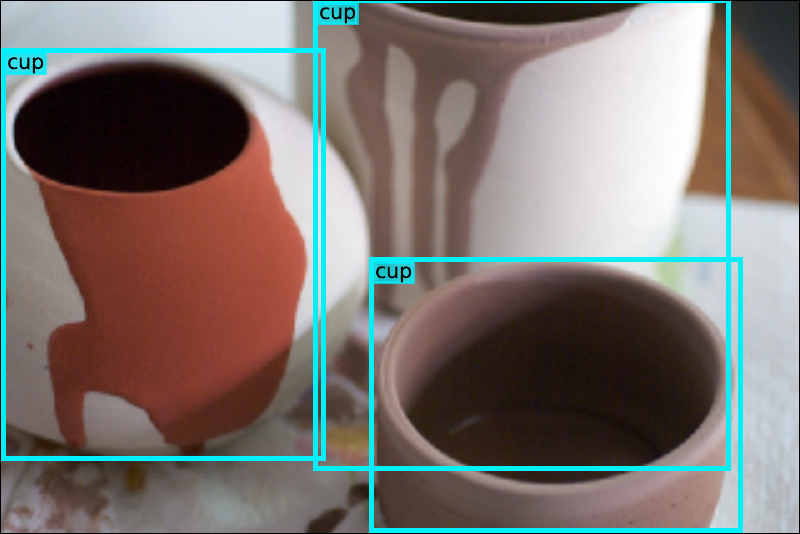}\includegraphics[height=2.6cm]{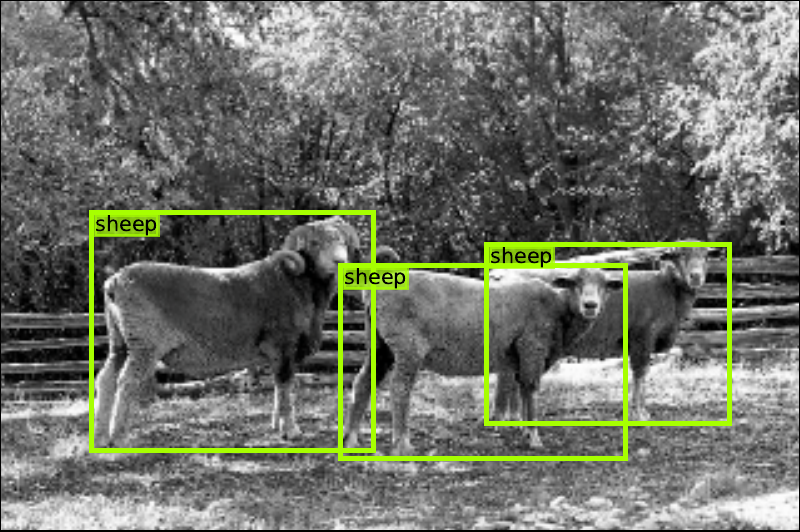}\includegraphics[height=2.6cm]{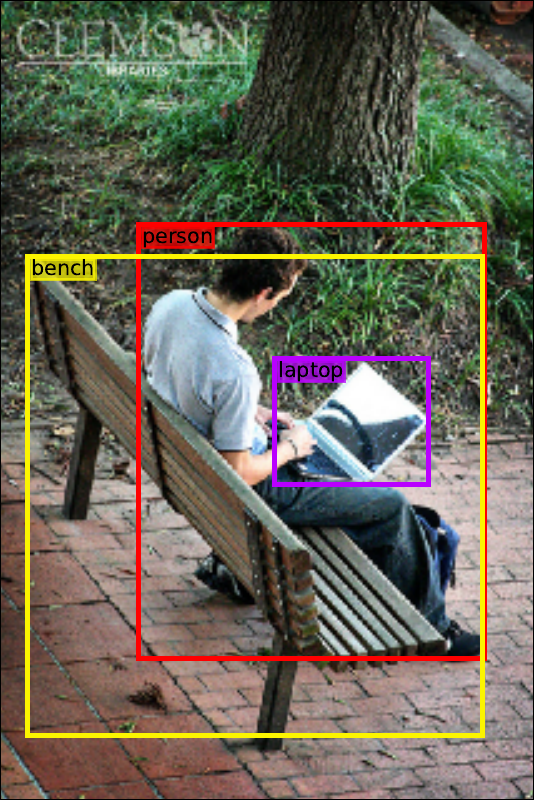}
\par\end{centering}
\begin{centering}
SM-NAS E3
\par\end{centering}
\caption{\label{fig:More-Qualitative-result-coco} Qualitative Results Comparison
on COCO dataset, tested on vanilla FPN with ResNet101 and our SM-NAS
E3.}
\end{figure*}

\begin{figure*}
\begin{centering}
\includegraphics[width=35mm]{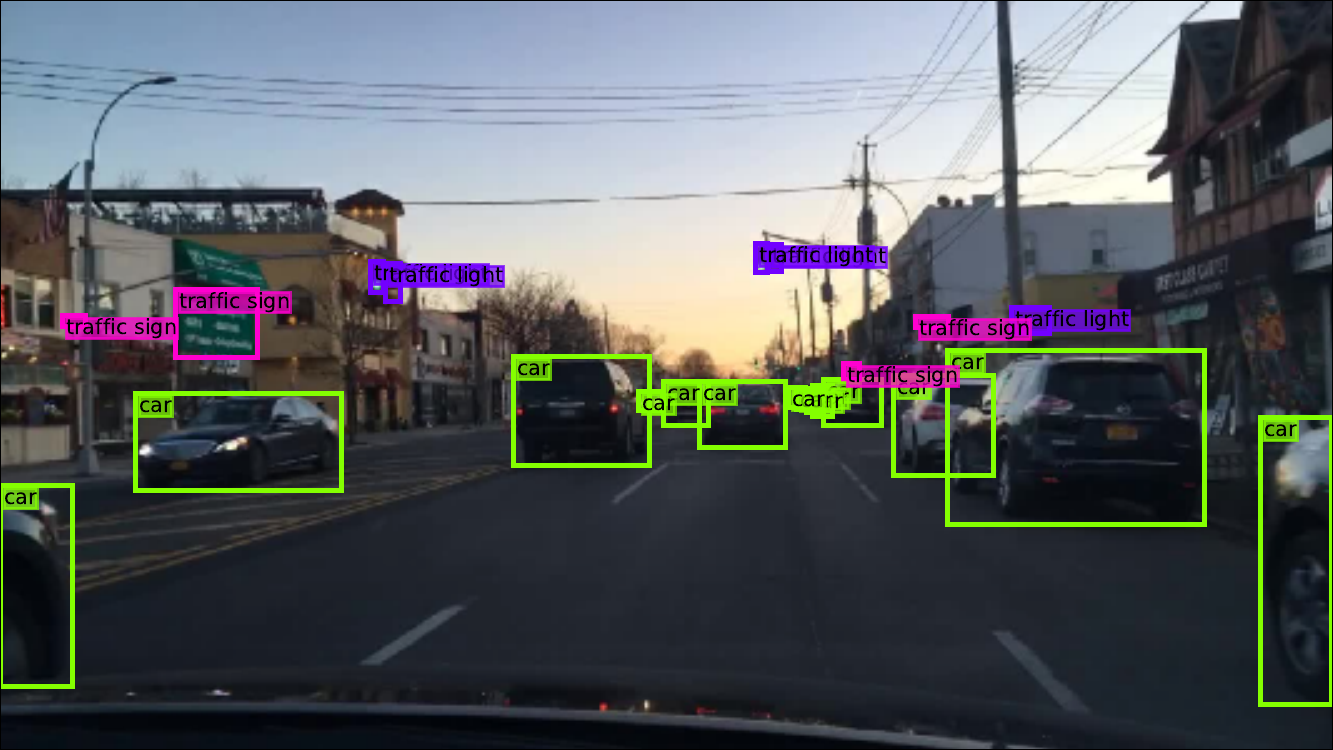}\includegraphics[width=35mm]{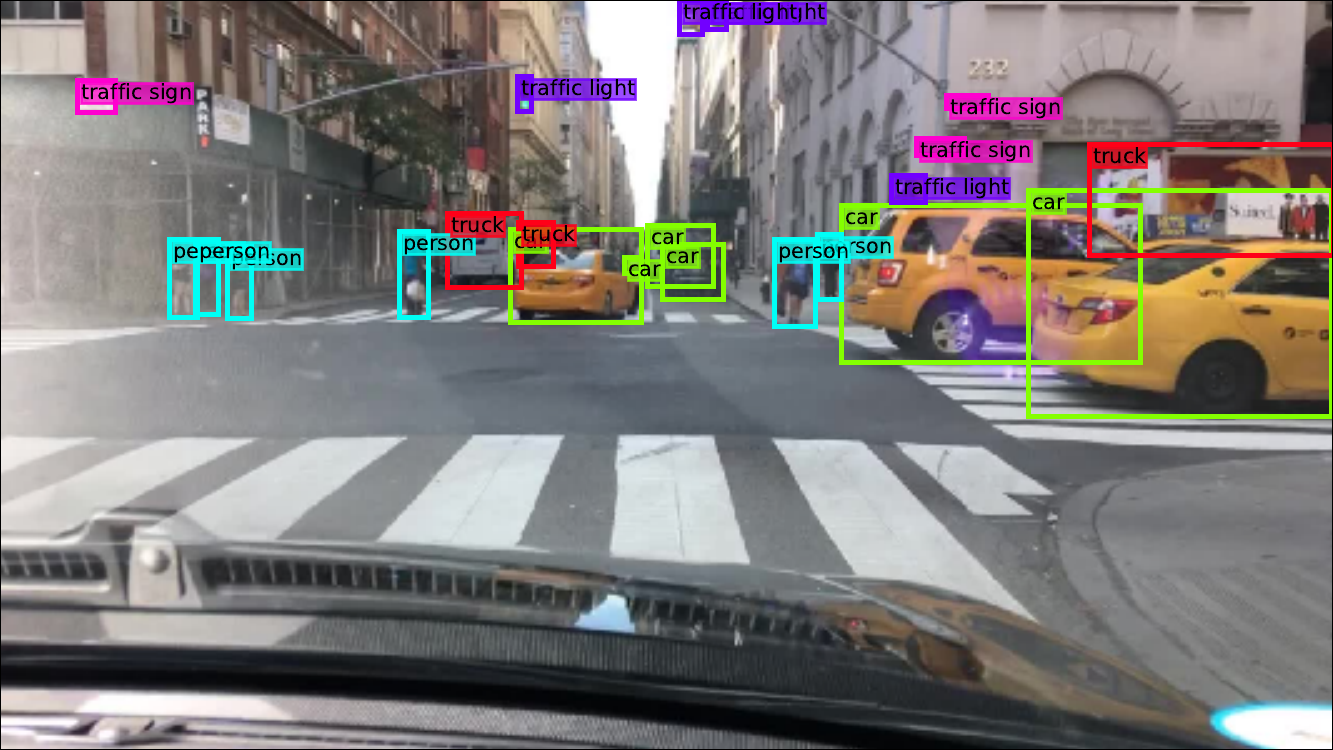}\includegraphics[width=35mm]{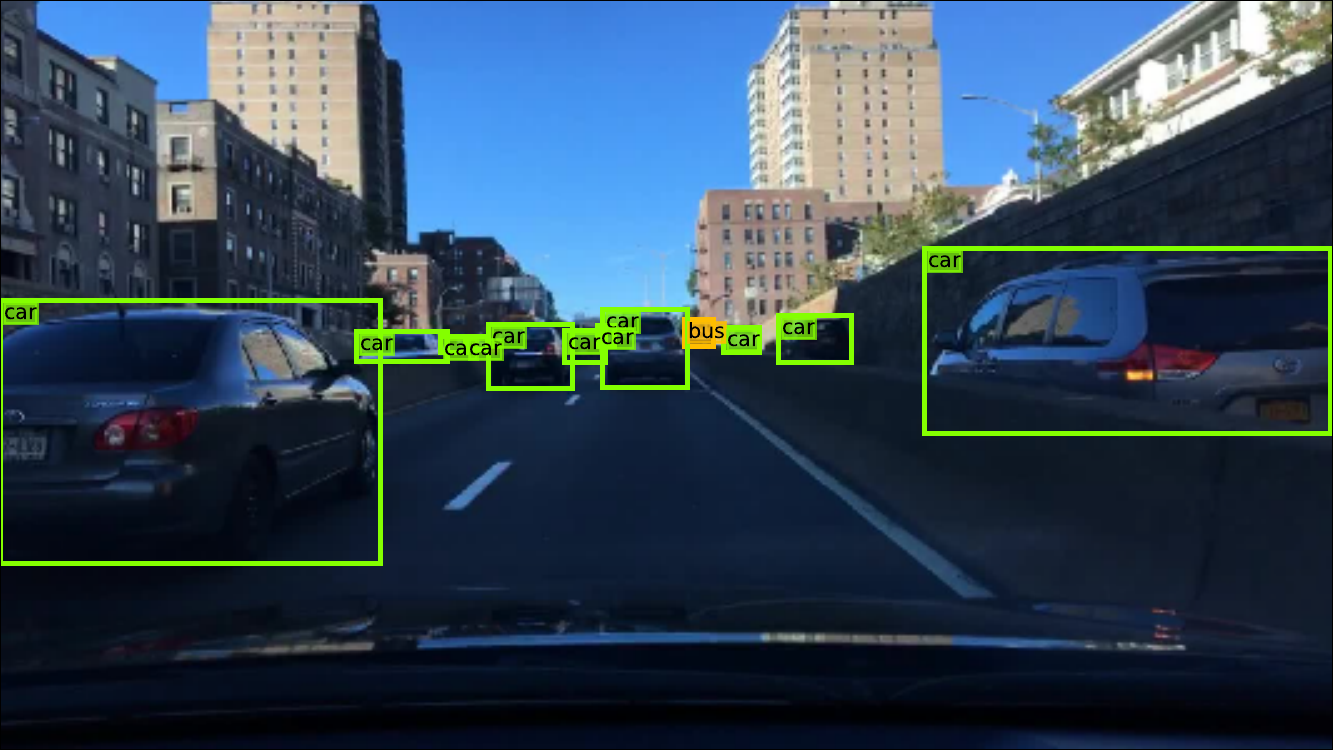}\includegraphics[width=35mm]{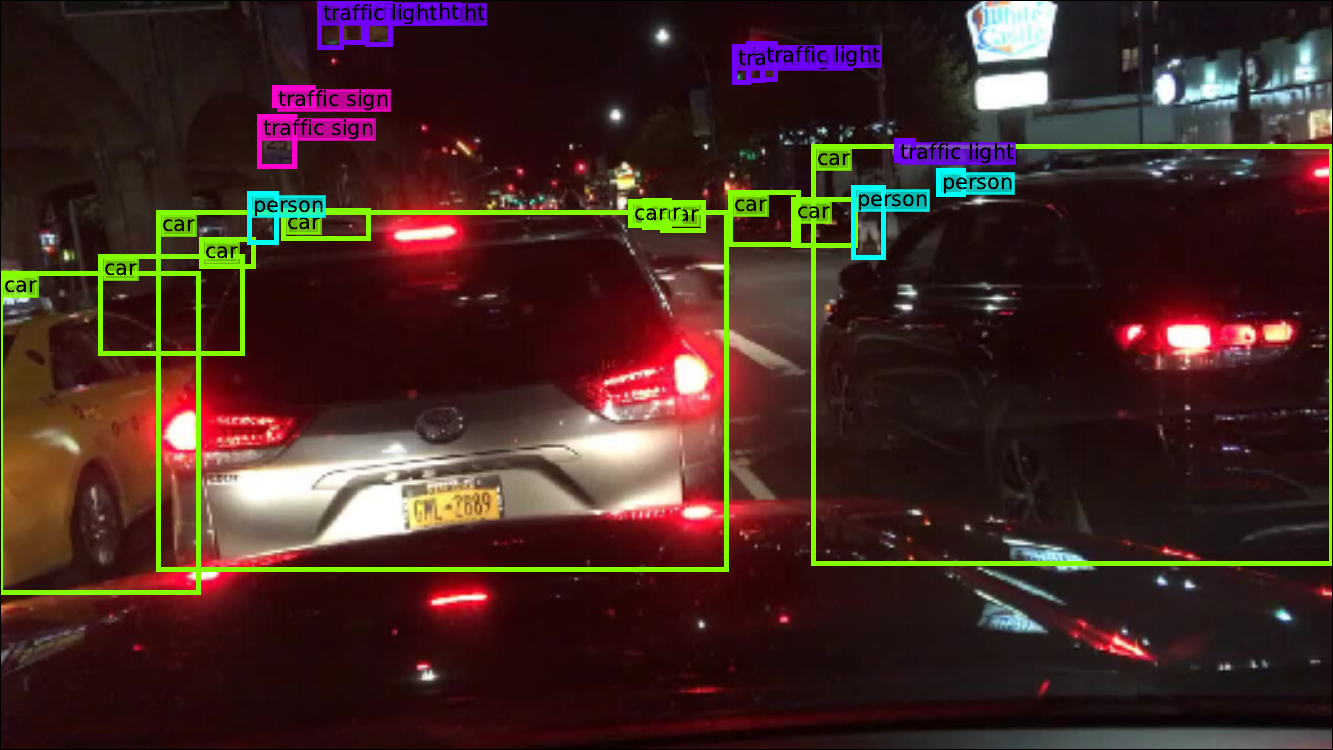}\includegraphics[width=35mm]{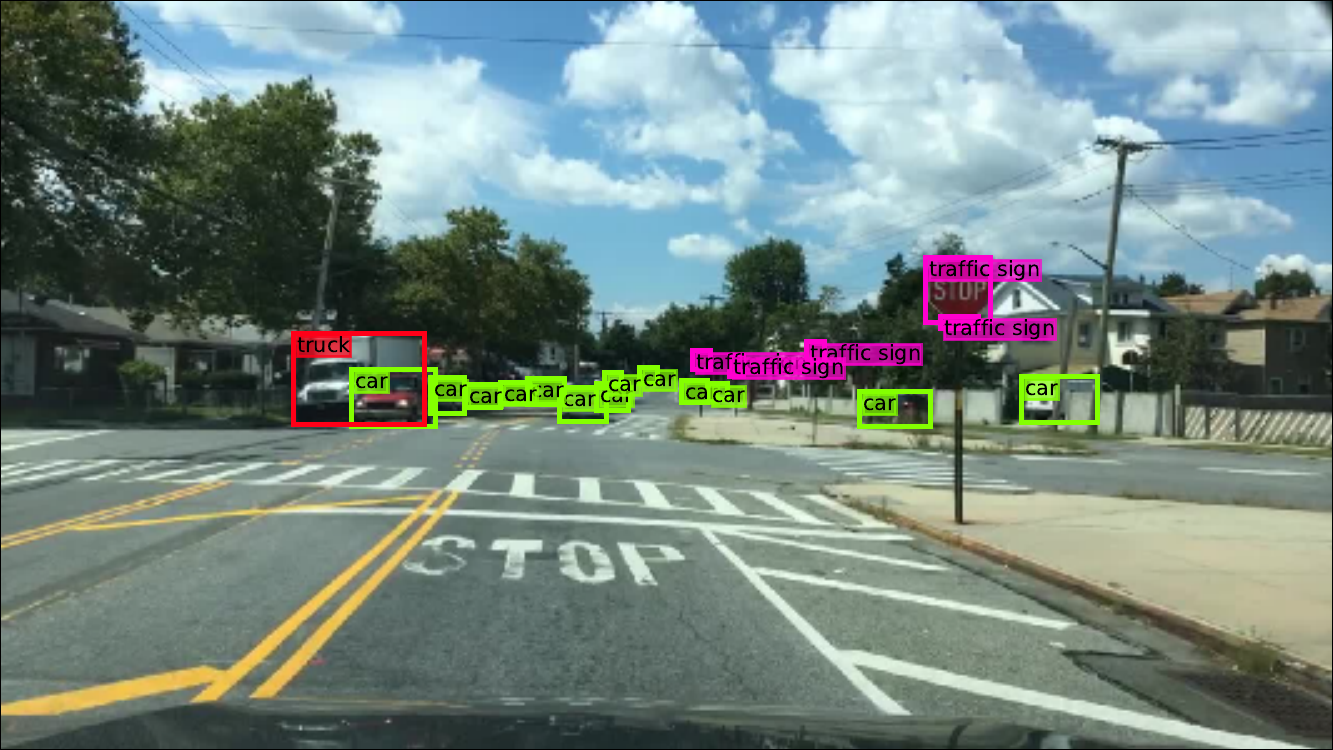}
\par\end{centering}
\begin{centering}
\includegraphics[width=35mm]{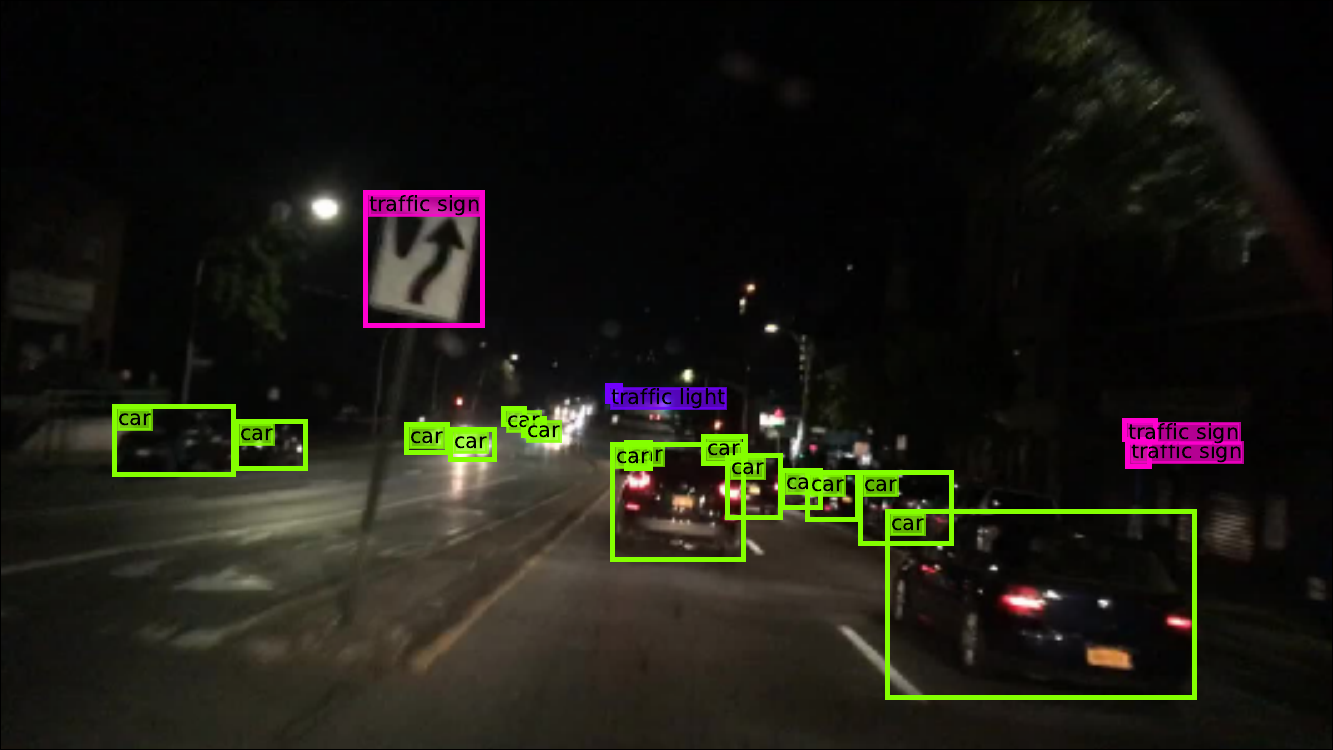}\includegraphics[width=35mm]{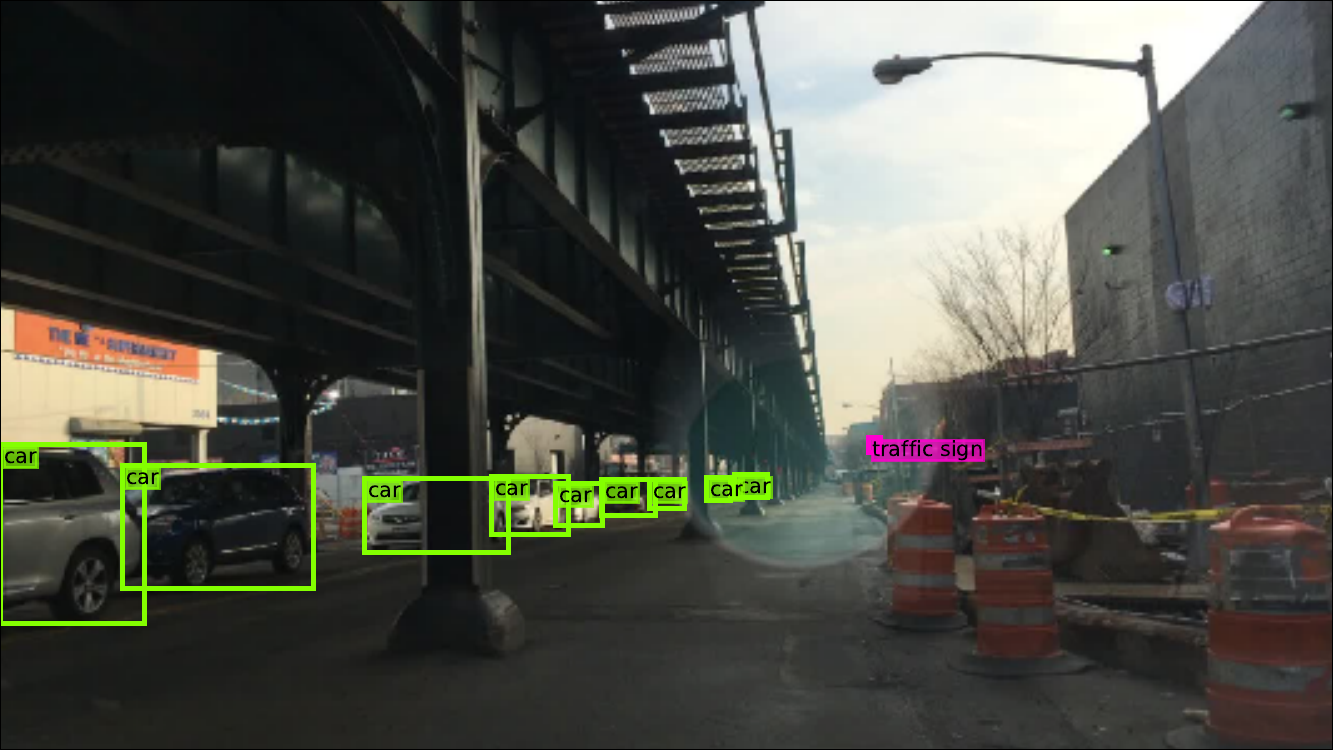}\includegraphics[width=35mm]{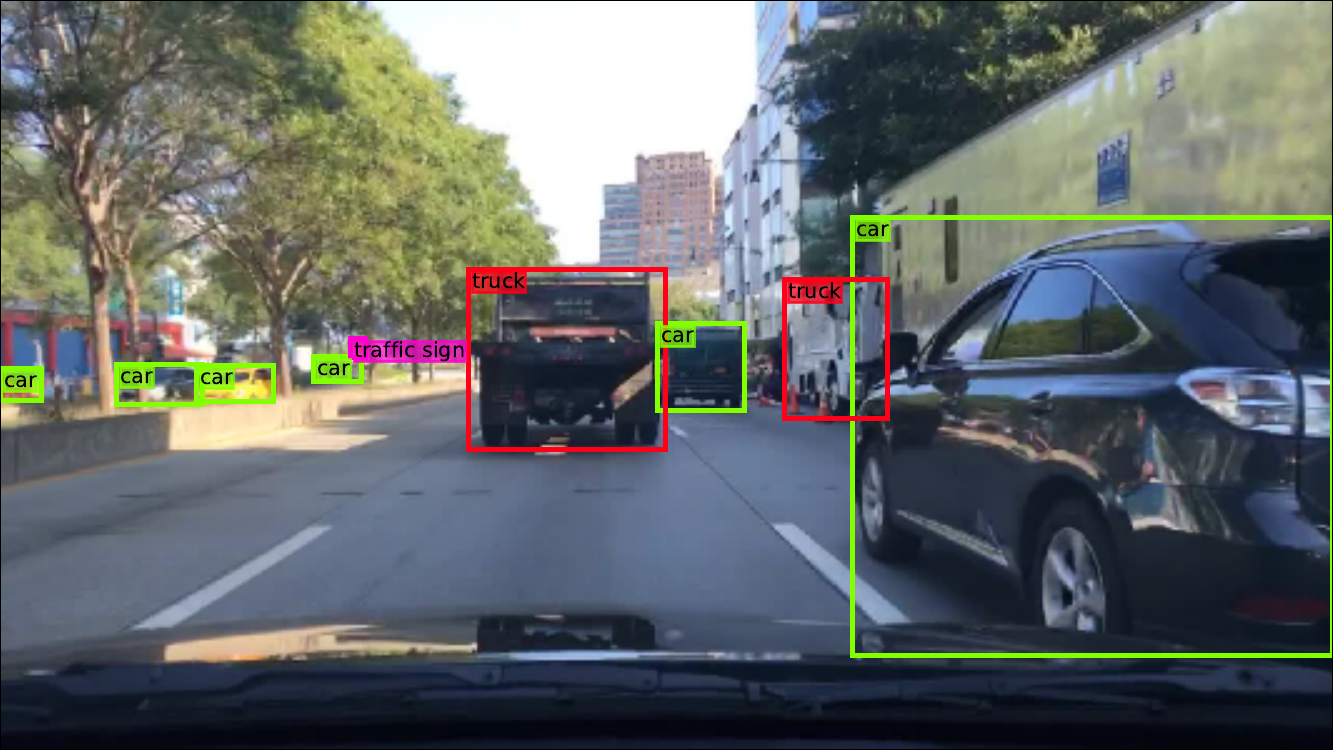}\includegraphics[width=35mm]{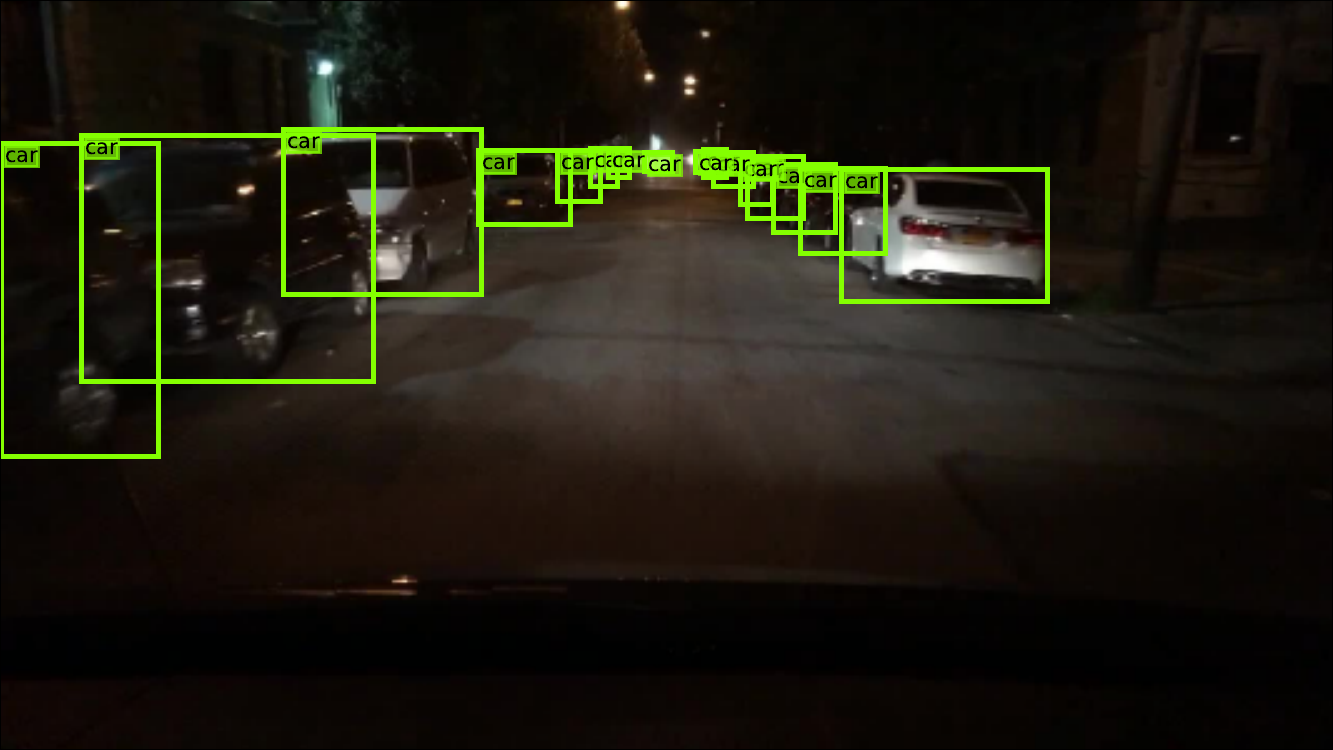}\includegraphics[width=35mm]{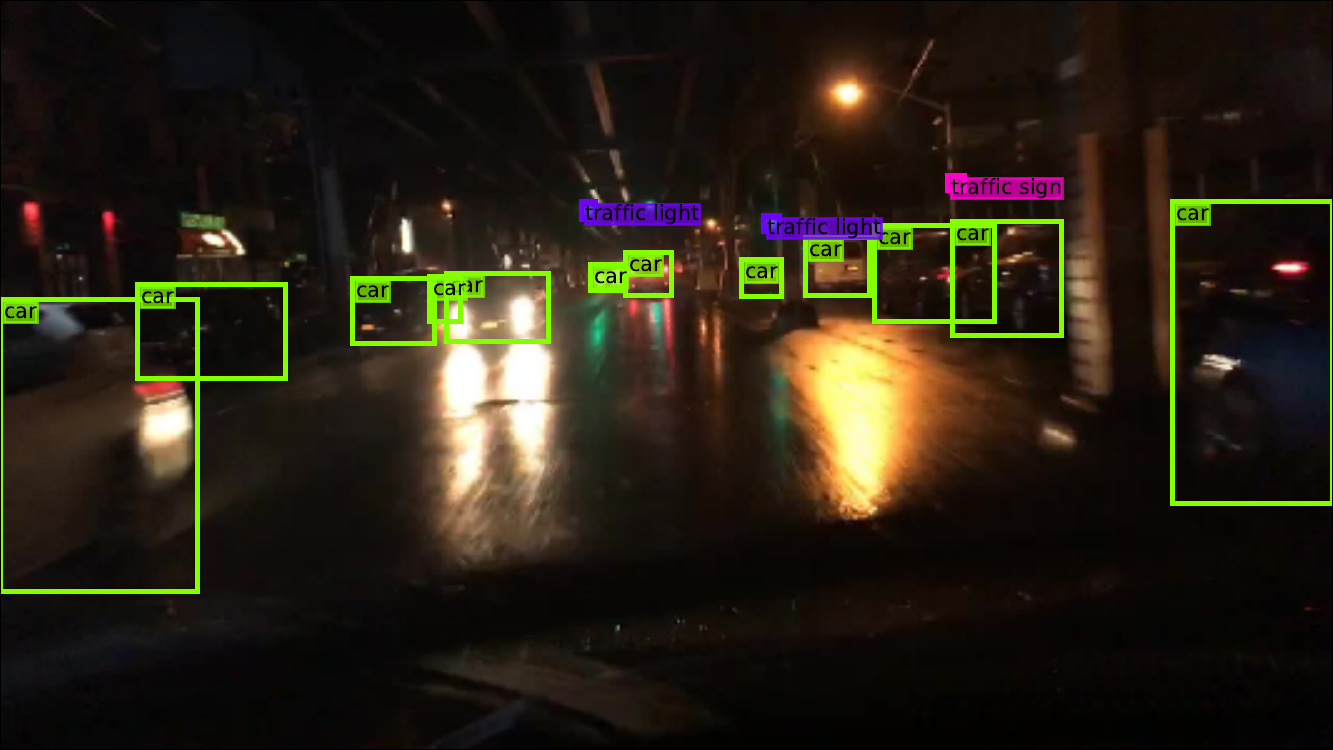}
\par\end{centering}
\caption{\label{fig:More-Qualitative-result-bdd}Qualitative results on BDD
dataset, tested on our SM-NAS E3.}
\end{figure*}

\begin{figure*}
\begin{centering}
\includegraphics[height=29mm]{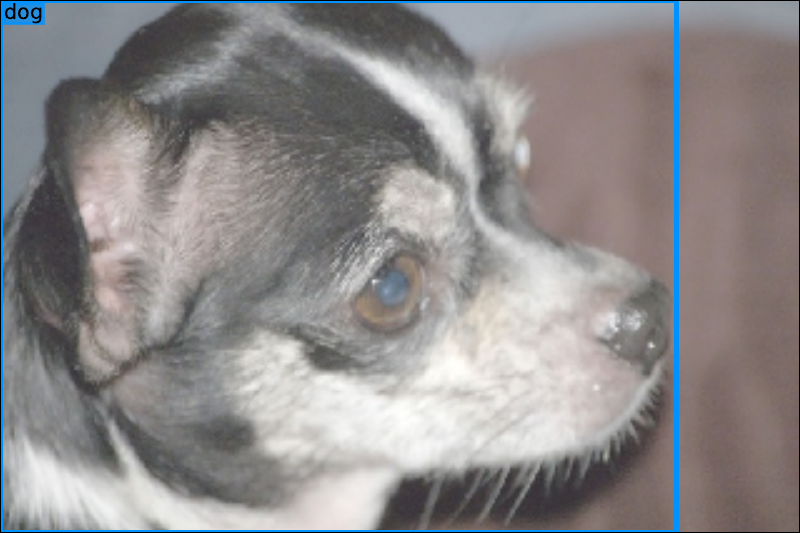}\includegraphics[height=29mm]{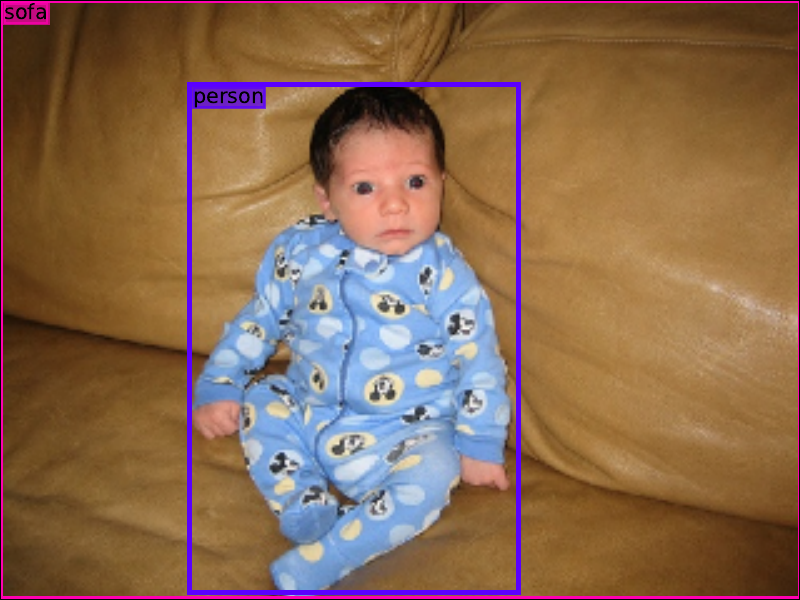}\includegraphics[height=29mm]{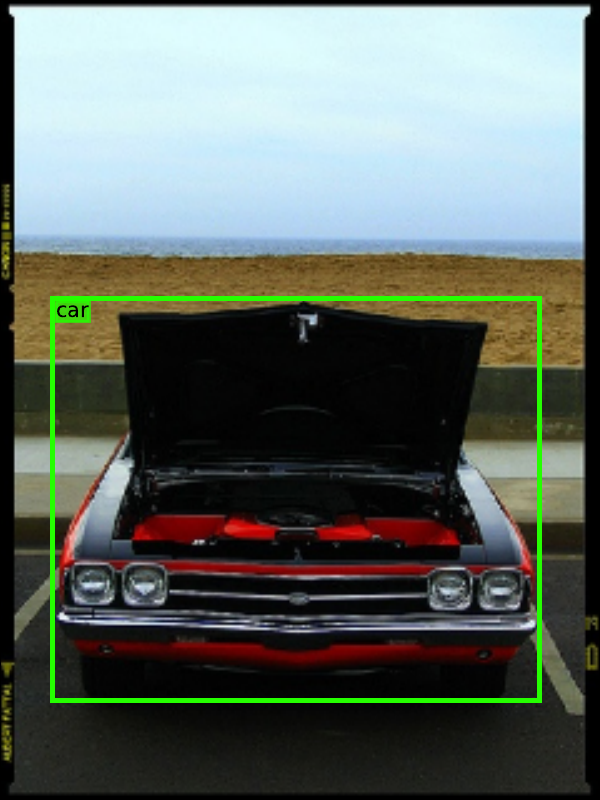}\includegraphics[height=29mm]{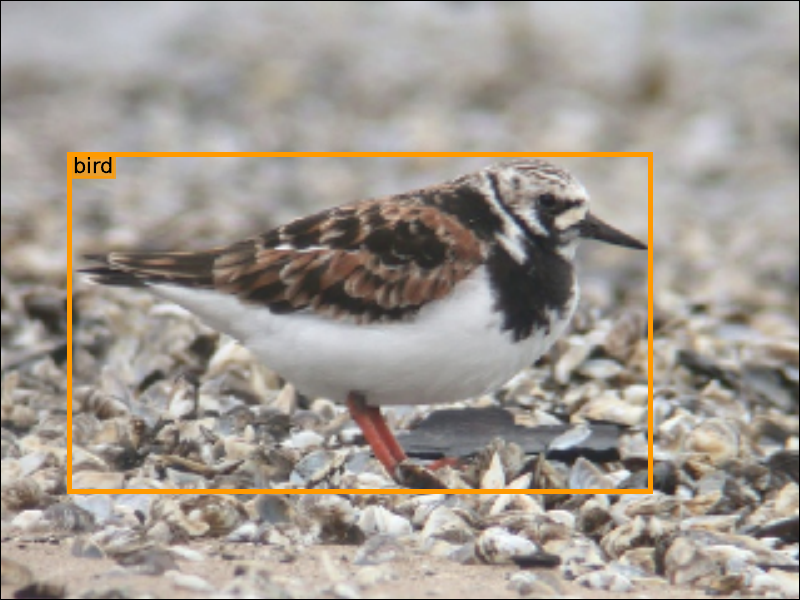}\includegraphics[height=29mm]{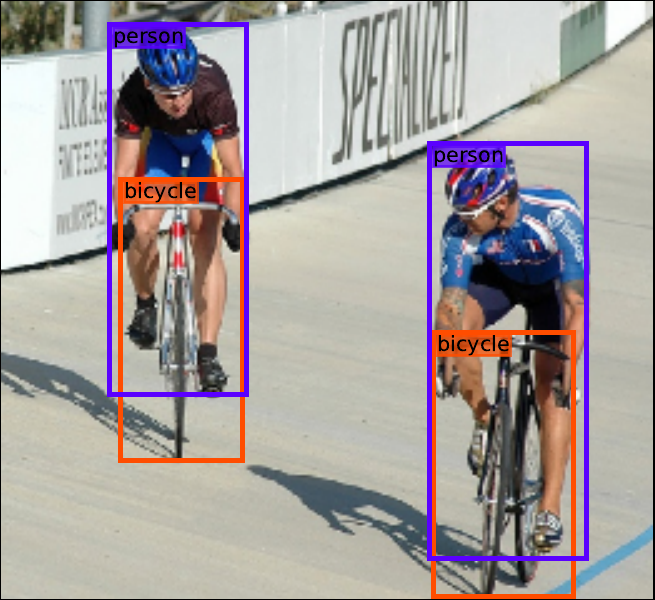}
\par\end{centering}
\begin{centering}
\includegraphics[height=29mm]{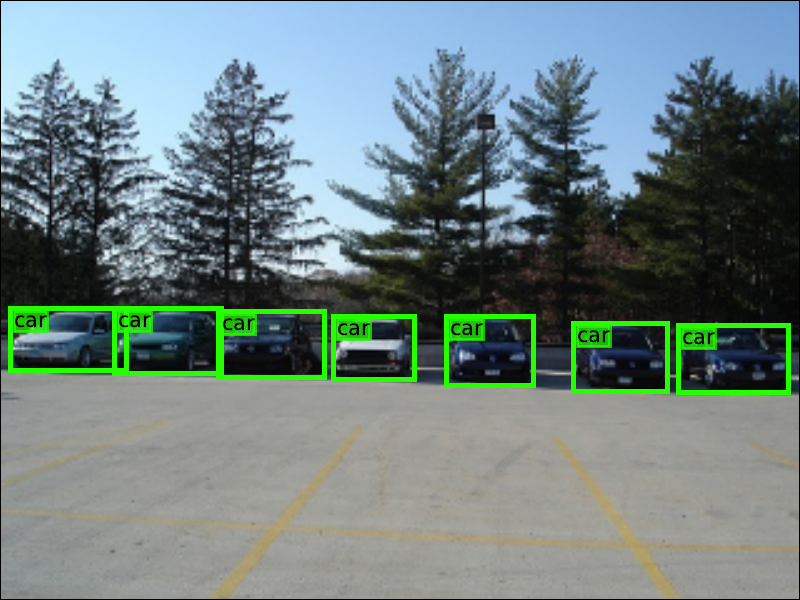}\includegraphics[height=29mm]{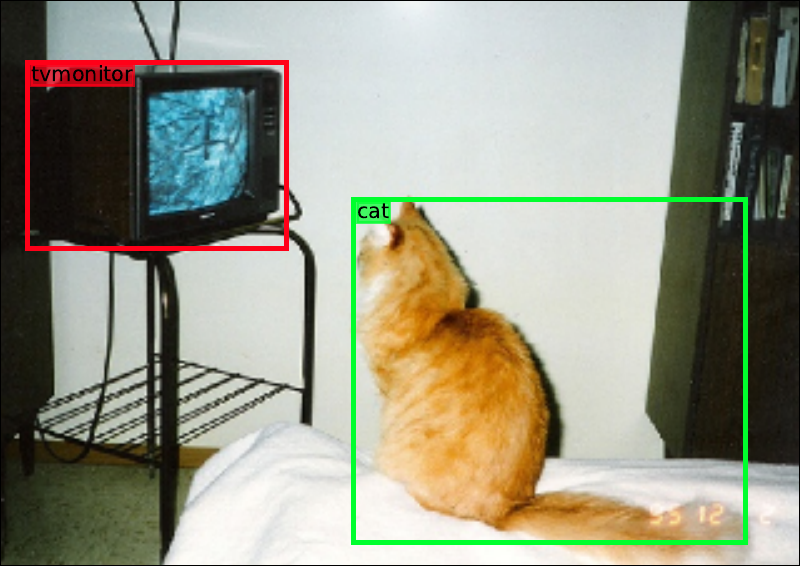}\includegraphics[height=29mm]{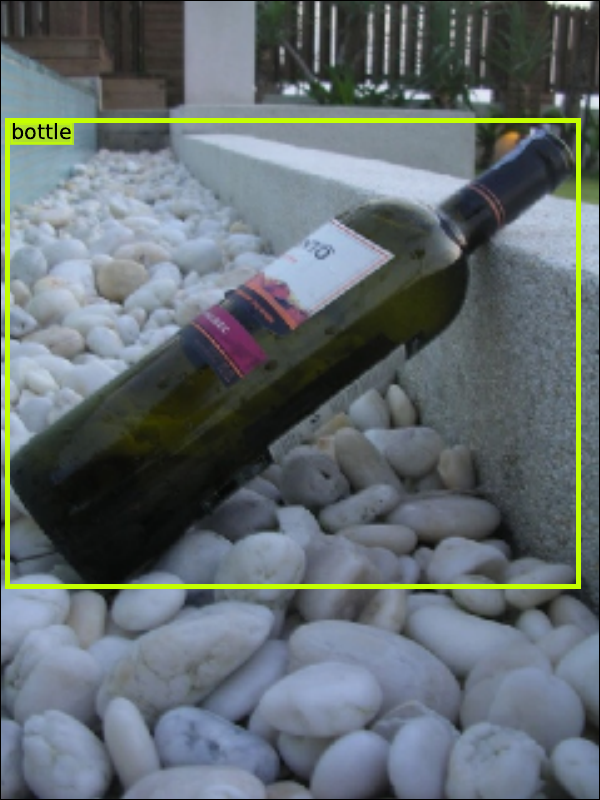}\includegraphics[height=29mm]{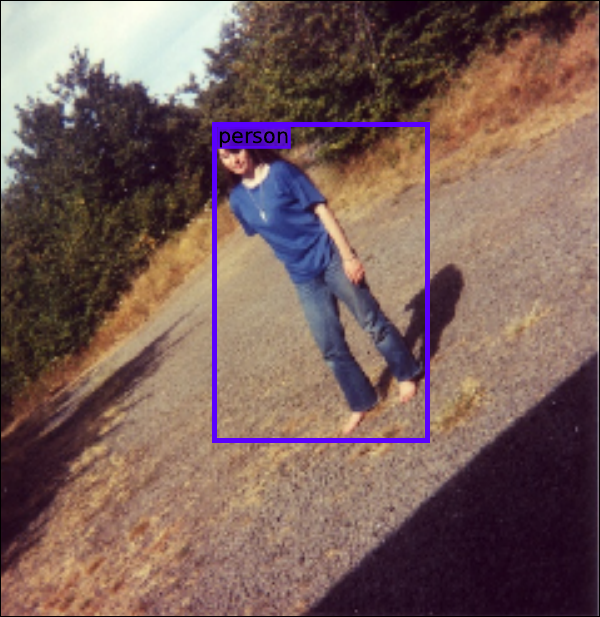}\includegraphics[height=29mm]{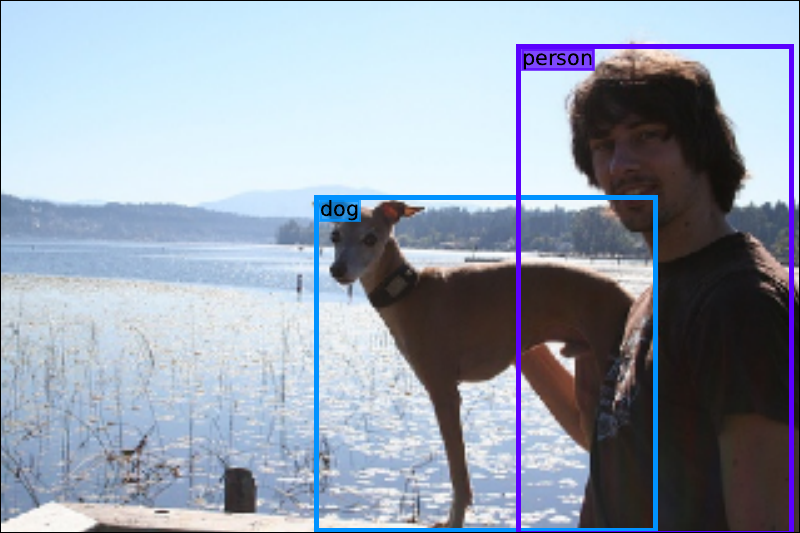}
\par\end{centering}
\caption{\label{fig:More-Qualitative-result-voc}Qualitative results on Pascal
VOC dataset, tested on our SM-NAS E3.}
\end{figure*}

More qualitative results comparison on multiple datasets: MSCOCO,
BDD, and Pascal VOC can be found in Figure \ref{fig:More-Qualitative-result-coco},
\ref{fig:More-Qualitative-result-bdd}, \ref{fig:More-Qualitative-result-voc}.
The SM-NAS E3 is our full model trained on all the three dataset.
The visualization threshold is 0.5. From Figure \ref{fig:More-Qualitative-result-coco},
our searched model is superior on the detection of objects with tiny-size,
occlusion, ambiguities to the baseline model FPN. From Figure \ref{fig:More-Qualitative-result-bdd},
it can be found that our E3 can detect very small cars. From Figure
\ref{fig:More-Qualitative-result-voc}, for a easier dataset Pascal,
our E3 performs also very well.

\end{document}